\documentclass[lettersize,journal]{IEEEtran}
\usepackage{amsmath,amsfonts}
\usepackage{algorithmic}
\usepackage{algorithm}
\usepackage{array}
\usepackage[caption=false,font=normalsize,labelfont=sf,textfont=sf]{subfig}
\usepackage{textcomp}
\usepackage{stfloats}
\usepackage{url}
\usepackage{verbatim}
\usepackage{graphicx}
\usepackage{cite}
\usepackage[switch]{lineno}
\usepackage{url}            
\usepackage[pagebackref,breaklinks,colorlinks,citecolor=darkerblue]{hyperref}  

\usepackage{booktabs}       
\usepackage{nicefrac}       
\usepackage{microtype}      
\usepackage{listings}

\usepackage{float}
\usepackage{multirow}
\usepackage{wrapfig}
\usepackage{xcolor}  
\usepackage{colortbl}
\usepackage{amssymb}
\usepackage[makeroom]{cancel}
\usepackage{cases}
\usepackage{orcidlink} 

\definecolor{gray}{rgb}{0.5,0.5,0.5}
\definecolor{gray94}{gray}{.92}
\definecolor{gray90}{gray}{.90}
\definecolor{gray85}{gray}{.85}

\newcommand{\ul}{\underline}

\definecolor{darkerblue}{RGB}{11,54,196}
\definecolor{DarkerGreen}{RGB}{21, 152, 56}
\definecolor{RoyalBlue}{RGB}{65,105,225}
\definecolor{YellowOrange}{RGB}{255,165,0}
\newcommand{\gray}[1]{\textcolor{gray}{#1}}
\newcommand{\red}[1]{\textcolor{red}{#1}}

\begin{document}
\title{GenURL: A General Framework for Unsupervised Representation Learning}

\author{
Siyuan Li$^{*,\orcidlink{0000-0001-6806-2468}}$%
,~\IEEEmembership{Student Member,~IEEE},
Zicheng Liu$^{*,\orcidlink{0000-0003-1106-2963}}$%
,~\IEEEmembership{Student Member,~IEEE},\\
Zelin Zang, Di Wu, Zhiyuan Chen, and Stan Z. Li$^{\dag,\orcidlink{0000-0002-2961-8096}}$%
,~\IEEEmembership{Fellow,~IEEE}
\thanks{Manuscript submitted October 18, 2022; accepted October 29, 2023. This work was supported in part by the National Key Research and Development Program of China under Grant 2022ZD0115100; in part by the National Natural Science Foundation of China under Project U21A20427; and in part by the Center of Synthetic Biology and Integrated Bioengineering, Westlake University, under Project WU2022A009.}
\thanks{Siyuan Li and Zicheng Liu contributed equally to this work. \dag~Stan Z. Li is the corresponding author.}
\thanks{The first two authors are from Zhejiang University, Hangzhou, 310000, China. Other authors are from the AI Division, School of Engineering, Westlake University, Hangzhou, 310030, China (e-mail: lisiyuan@westlake.edu.cn; liuzicheng@westlake.edu.cn;~zangzelin@westlake.edu.cn;~wudi@westlake.edu. cn;~chengzhiyuan@westlake.edu.cn;~stan.zq.li@westlake.edu.cn).}
}

\markboth{IEEE Transactions on Neural Networks AND
Learning Systems}%
{Shell \MakeLowercase{\textit{et al.}}: A Sample Article Using IEEEtran.cls for IEEE Journals}


\maketitle

\begin{abstract}

Unsupervised representation learning (URL), which learns compact embeddings of high-dimensional data without supervision, has made remarkable progress recently. 
However, the development of URLs for different requirements is independent, which limits the generalization of the algorithms, especially prohibitive as the number of tasks grows.
For example, dimension reduction methods, t-SNE, and UMAP optimize pair-wise data relationships by preserving the global geometric structure, while self-supervised learning, SimCLR, and BYOL focus on mining the local statistics of instances under specific augmentations.
To address this dilemma, we summarize and propose a unified similarity-based URL framework, GenURL, which can smoothly adapt to various URL tasks.
In this paper, we regard URL tasks as different implicit constraints on the data geometric structure that help to seek optimal low-dimensional representations that boil down to data structural modeling (DSM) and low-dimensional transformation (LDT). 
Specifically, DMS provides a structure-based submodule to describe the global structures, and LDT learns compact low-dimensional embeddings with given pretext tasks.
Moreover, an objective function, General Kullback-Leibler divergence (GKL), is proposed to connect DMS and LDT naturally.
Comprehensive experiments demonstrate that GenURL achieves consistent state-of-the-art performance in self-supervised visual learning, unsupervised knowledge distillation (KD), graph embeddings (GE), and dimension reduction.
The code is available at \url{https://github.com/Westlake-AI/openmixup}.

\end{abstract}

\begin{IEEEkeywords}
Contrastive learning (CL), dimension reduction (DR), graph embedding (GE), knowledge distillation (KD), self-supervised learning (SSL)
\end{IEEEkeywords}

\section{Introduction}

\IEEEPARstart{L}{}earning low-dimensional representations from complex data without human supervision, \textit{i.e.}, unsupervised representation learning (URL), is a long-standing problem. As the high-dimensional data is usually highly redundant and non-Euclidean, a widespread assumption is that data lies in a low-dimensional ambient space.
{However, URL algorithms under different tasks and data structures are designed independently of each other, yet they have the same ultimate goal of finding the desired embedding space.}

\begin{figure}
    \centering
    \includegraphics[width=0.96\linewidth]{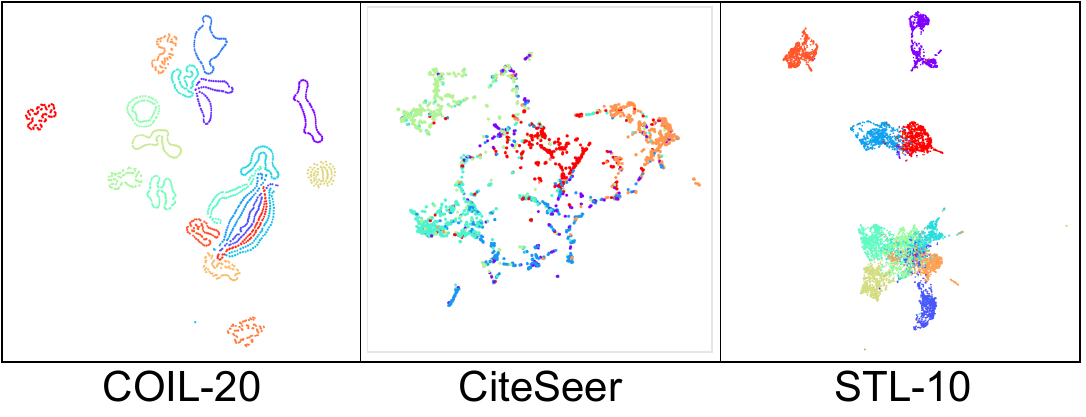}
    \vspace{-1.0em}
    \caption{Illustration of various empirical structures of high-dimensional data. We encode COIL-20~\cite{1996cOIL20}, CiteSeer~\cite{1998citeseer}, and STL-10~\cite{2011stl10} to 2-dim, 128-dim, and 512-dim by GenURL (128-dim and 512-dim latent spaces are then visualized by UMAP~\cite{2018UMAP} in 2-dim). \textit{Left}: we preserve local geometric structures of the circle manifolds in COIL-20 in the DR task. \textit{Middle}: the topological and geometric structures of citation networks in CiteSeer are encoded in the GE task. \textit{Right}: with the instance discriminative proxy task, we learn a discriminative representation in the validation dataset of STL-10.}
    \label{fig:introduction}
    \vspace{-1.0em}
\end{figure}

URL studies are now broadly categorized into three types of applications.
(i) Dimension reduction (DR) and graph embedding (GE) algorithms aim to encode non-Euclidean input data into a latent space $Z$ plainly \textit{without any prior knowledge} of the related domains, as shown in Figure~\ref{fig:introduction} left and middle. (ii) Complementary to DR and GE, another popular path of URL focuses on data-specific augmentations, such as image crop, which leads to a clustering structure and learns discriminative representations, as illustrated in Figure~\ref{fig:introduction} right. (iii) In addition, knowledge distillation (KD) is another approach that can be regarded as an implicit URL method transferring the knowledge from the pretrained model to enhance the student model unsupervised instead of considering geometric structure or special prior information of the target data.
Specifically, from the perspective of algorithmic bias, these representative URL algorithms are grouped into two classes: global structure-oriented \textit{e.g.,} t-SNE and UMAP, \textit{etc.}, and individual augmentation-oriented \textit{e.g.,} SimCLR, MoCo, \textit{etc.}, respectively.
It is a fact that these two independent algorithms are designed to excel in their respective areas of applicability.
\textbf{There is thus a natural question if the intrinsic representation of data is determined by both the global data structure and data-specific prior assumptions in a unified framework.}

\textbf{This work: A general framework of unsupervised representation learning.}
Based on the above URL methods, developing an effective and unified URL framework adaptive to various scenarios is a new trend in the community~\cite{mu2021slip, baevski2022data2vec}. 
In this work, we consider both the global structure and local discriminative statistics and thus reformulate the URL problem as a non-Euclidean data embedding problem that encodes the structure and content parallelly in a compact low-dimensional space. 
For instance, we introduce an effective and general framework of unsupervised representation learning called GenURL, containing two steps: data structural modeling (DSM) and low-dimensional Transformation (LDT).
To model the global structures, we combine the graph distance calculated on both the \textit{raw feature space} and \textit{predefined graphs}, which define the task-specific knowledge, \textit{e.g.}, the provided graph $\mathcal{G}$ for GE tasks and the proxy task of SSL tasks.
To learn the embedding according to data structures, we define a similarity between each sample pair and corresponding latent representations based on the pair-wise distances and design robust loss functions to optimize the encoder $f_{\theta}$.
Unlike previous DR and GE methods, the proposed GenURL can import extra prior knowledge and is robust to highly redundant data; additionally, different from GE and SSL methods, GenURL is agnostic to network structures and predefined proxy tasks. 
Extensive experiments conducted on benchmarks of four URL tasks (self-supervised visual presentation learning, unsupervised knowledge distillation, graph embedding, and dimension reduction) show that GenURL achieves state-of-the-art performance. We further analyze the relationship between empirical data structures in various tasks and the loss functions and hyper-parameters in GenURL.
In short, this paper makes three contributions:
\begin{itemize}
    \item We propose a unified and general URL framework (GenURL) that encodes the global structure and local discriminative statistics of input data into a compact latent space parallelly.
    \item We discuss two types of embedding problems based on whether the input distance is well-defined and propose a novel objective function named General Kullback-Leibler divergence (GKL) to connect global and local efficiently.
    \item We adapt GenURL to various scenarios to verify the effectiveness and explain the relationship between various tasks through extensive experiments.
\end{itemize}

\section{Related Work}
\label{ch2:RelatedWork}
\paragraph{Dimension reduction}
Adopting the manifold assumption in DR, which assumes data lie on a low-dimensional manifold immersed in the high-dimensional space, most DR methods try to preserve intrinsic geometric properties of data~\cite{sci2000Isomap, sci2000LLE, siam2004LTSA, 2003Hessian, icml2020TopoAE, melacci2012unsupervised}. Another practical branch of DR introduced by t-SNE~\cite{jmlr2008tSNE} and UMAP~\cite{2018UMAP} optimizes the pair-wise similarities between latent and input spaces. More recently, deep manifold learning methods can learn more complex manifolds and can be transferred to unseen data by using neural networks. Parametric t-SNE (P-SNE)~\cite{jmlr2009PSNE}, Parametric UMAP (P-UMAP)~\cite{2021pUMAP}, DMT~\cite{2021dmt}, and its variant~\cite{ecml2021invML} are proposed directly based on t-SNE and UMAP. {However, current DR methods model desired data structures only relying on the geometry of input space and might fail with highly redundant data, such as natural images.}

\paragraph{Graph Embedding}
GE transfers graph data into a low-dimensional and continuous feature space while preserving most graph structures and topological and geometric structures, such as vertex content. Most early methods are model-free, which contains four categories: Laplacian eigenmaps-based~\cite{newman_finding_2006}, local similarity-based~\cite{icml2014SkipGram, kdd2014DeepWalk,zhang2020unsupervised}, matrix factorization-based~\cite{aaai2018community}, and nonparametric Bayesian modeling-based methods~\cite{aaai2018bojchevski_bayesian}. Recently, model-based methods~\cite{kipf2016GVAE, 2019AGC} utilize graph convolutional networks (GCN)~\cite{2016PubMed} or graph auto-encoders~\cite{2020ARGA, kdd2020AGE} to learn both graph structures and feature information. More recently, some methods~\cite{Zang2021DMAGE} take both the geometric and topological structures into consideration. 
This type of learning, which relies exclusively on the graph structure, eventually leads to the problem of homogenization of representations, so in addition to conveying information through the structure, the nodes themselves need to be bounded by independent a priori information.

\paragraph{Self-supervised Visual Representation Learning}
Early SSL methods design hand-crafted pretext tasks \cite{iccv2015relativeloc, eccv2016coloring, iclr2018rotation,he2022analyzing}, which rely on somewhat ad-hoc heuristics and have limited abilities to capture practically useful representations. Another popular form is clustering-based methods \cite{eccv2018deepcluster, cvpr2020odc, iclr2020sela, nips2020swav} learning discriminative representation by offline or online clustering. Recently, contrastive learning (CL) \cite{eccv2018npid, cvpr2020moco, 2020simclr, nips2020byol}, which discriminates positive pairs against negative pairs, achieved state-of-the-art performance in various vision tasks. Different mechanisms \cite{nips2020byol, cvpr2021simsiam, icml2021barlow, eccv2022dlme} are proposed to prevent trivial solutions in CL to learn useful representations. To fully utilize negative samples, \cite{nips2020debiased, nips2020mochi, 2021iclrHCL, li2021samix} explore hard samples in the momentum memory bank. Meanwhile, some efforts have been made on top of contrastive methods to improve pre-training quality for specific downstream tasks~\cite{iccv2021detco, xiao2021region, cvpr2021casting, wu2021align}. 
More recently, with the introduction of the Vision Transformers~\cite{dosovitskiy2020image,iccv2021Swin}, masked image modeling (MIM)~\cite{bao2021beit, he2021masked, xie2021simmim, Li2022A2MIM} achieved state-of-the-art performance based on Transformers, which randomly mask out patches in the input image and predict the masked patches with decoder.
While SSL can extract highly redundant data features, the loss of geometric structure constraints is an inability to model a priori semantic relationships between clusters in a global context, such as temporal evolutionary order.

\paragraph{Knowledge Distillation}
KD was first proposed by \cite{nips2014kd}, which aims to transfer knowledge from trained neural networks to a smaller one without losing too much generalization power. There are three types of existing KD methods: response-based~\cite{nips2014kd, cvpr2018dml, cvpr2020noisy} and feature-based~\cite{iclr2017attkd} methods require labels to utilize intermediate-level supervision from the teacher model. Relation-based KD~\cite{cvpr2017gift, eccv2018pkt, cvpr2019rkd, iccv2019sp, iclr2021seed} methods explore the relationships between different layers or data sample pairs which can work without supervision and extend to self-supervised settings~\cite{eccv2020SSKD, iclr2020CRD, iclr2021seed}. However, The performance of replicating the teacher model alone is unsatisfactory, and considering both the data structure and the a priori assumptions is a critical step in improving the efficiency of transfer learning. Our method handles the KD task as a special type of DR task without label supervision.

We summarize and analyze the above URL approaches and propose GenURL, a framework that successfully links the two important elements of URLs, data geometry and task hypotheses, based on generalized similarity. GenURL not only takes into account the data geometries that are the focus of DE and GE but also introduces task-relevant data and model hypotheses from CL and KD to enhance the overall quality of the representations and thus improve the downstream tasks. performance of the downstream tasks.

\section{Method}
\label{sec:Method}

\begin{figure*}[t]
    \centering
    \includegraphics[width=0.95\linewidth]{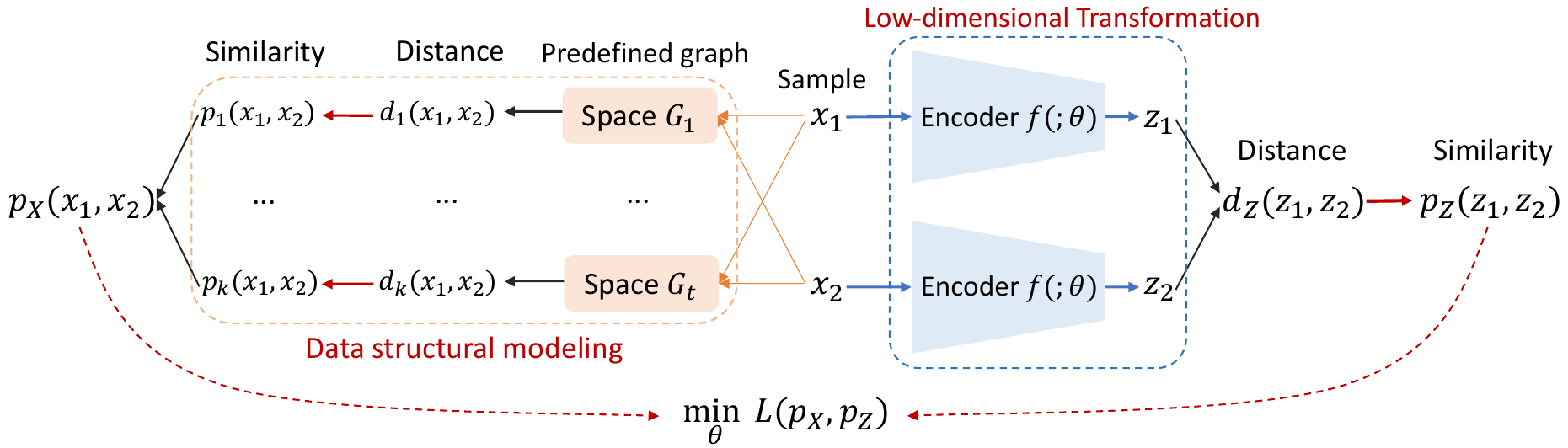}
    \caption{Illustration of GenURL. The data structures are first modeled as similarity $P_{X}$ by calculating the graph distance on each predefined graph. Then, the low-dimensional transformation mapping $f_{\theta}$ is optimized by minimizing $\mathcal{L}$ based on the fixed $P_{X}$.}
    \label{fig:pipeline}
\end{figure*}

\begin{figure}[b]
    \centering
    \includegraphics[width=0.98\linewidth]{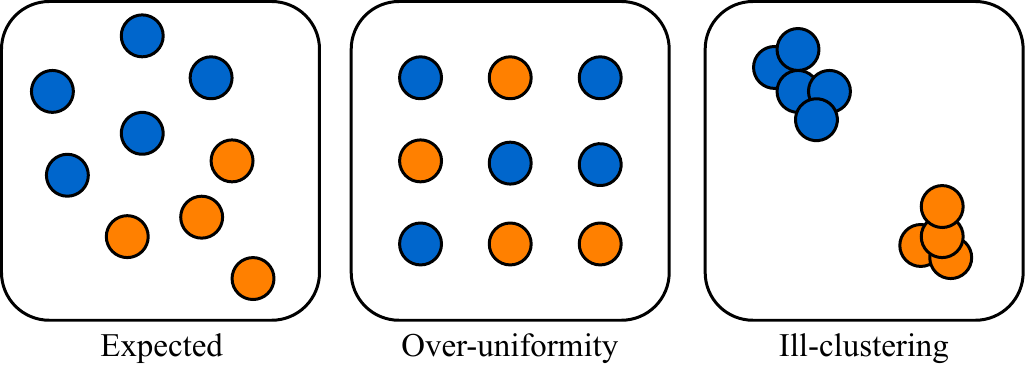}
    \vspace{-0.25em}
    \caption{Illustration of two typical issues in unsupervised learning tasks: over-uniformity and ill-clustering.}
    \label{fig:issues}
\end{figure}

\subsection{Preliminaries}
\label{subsec:pre}
The general goal of URL is identical across various URL tasks:
Given a finite set of samples $X=[x_1,...,x_n]\in \mathbb{R}^{n\times D}$, we seek a continuous mapping, $f_{\theta}(x):\mathbb{R}^{D}\rightarrow \mathbb{R}^{d}$, where $d \ll D$, that transfers data into a compact latent space $Z$ while preserving essential structures of $X$ to facilitate most downstream tasks~\cite{cvpr2016resnet, iccv2017maskrcnn, 2019AGC}.
Generally, the intrinsic structures of data are determined by \textit{both the input data and task-specific assumptions}.
However, most research in URL has focused on individual data modalities or specific tasks, resulting in specific designs and different learning objectives. We take the following four widely used URL tasks as examples.

Although there are various task-specific assumptions in URL tasks, we may summarize them as a data embedding problem based on the manifold assumption.
Assuming the data $X$ is constrained on a compact low-dimensional manifold $\mathcal{M}$, a neighborhood system for each sample $x_i$ is defined as $\mathcal{N}_i\in \mathbb{R}^{d}$, and $\mathcal{M}$ is supported on the mixture of these neighborhood systems, $\mathcal{M} = \cup_{i=1}^{n}\mathcal{N}_i$. 
The discriminative property of $\mathcal{N}$ facilitates various downstream tasks.
Since different neighborhood systems are disjoint, we use an adjacency matrix $A$ to represent neighborhood systems: $A_{ij}=1$ indicates $x_i$ and $x_j$ are in the same neighborhood system, or $A_{ij}=0$. 
$A$ can be built by an undirected graph $\mathcal{G}=(X,E)$ provided by a specific URL task (discussed in Sec.~\ref{sec:instantiation}). 
Local (neighborhood systems) and global (manifold) structures are two essential properties for learning good representations: local geometries describe the discriminative features of instances, while relationships between neighborhood systems reflect the global view of topological structures. 
Based on $\mathcal{M}$, the similarity between two non-adjacent samples within each $\mathcal{N}_i$ can be approximated by the shortest-path distance.
Since most URL algorithms are designed for specific tasks or data, the following two typical issues arise:

\textbf{Over-uniformity.}\quad
Without the global structure, the optimal solution for constructing $\mathcal{N}$s is to place each $x$ evenly in the embedding space, as shown in the middle of Fig~\ref{fig:issues}. 
In other words, the decision boundaries are maximized to the discrimination between $x$. 
However, the manifold structure is then damaged and unorganized in this case, which means the learned representations can not be generalized to other downstream tasks~\cite{wang2020understanding}, such as data visualization.

\textbf{Ill-clustering.}\quad
The converse is also true; if we focus excessively on the global structure and ignore instance differences, a local collapse will occur. 
The result is shown on the right of Fig~\ref{fig:issues}, also called local homogenization.
A classical case in the node classification task of graph data is the over-smoothing issue~\cite{cai2020note}: global-based message passing makes the connected nodes ultimately non-distinguishable.

Therefore, there is a question worth thinking about \textit{whether we can solve both of these issues at the same time in a mutually constraining way.}
Motivated by this, we propose a general and unified framework for URL that can be adapted to various URL tasks effectively.

\subsection{Similarity-preserving Framework}
\label{subsec:framework}
Given a set of $m$ empirical graphs $\mathcal{G}=\{G_t\}_{t=1}^m$ defined on the data $X$, where $\mathcal{G}_t=(X,d_{t})$, we calculate the pair-wise distance $d_{X,t}$ based on $X$ and $\mathcal{G}_t$ to model the empirical data structures. To capture the local geometry defined in $\mathcal{G}_t$ and eliminate the scale factor between different distances, we adopt the \textit{similarly} defined in $[0,1]$ by converting the pair-wise distance $d$ to the similarity with a long-tailed t-distribution kernel function $\kappa(.)$,
\begin{equation}
    \kappa (d,\nu) = \sqrt{2\pi} \cdot\frac{\Gamma\left(\frac{\nu+1}{2}\right)} {\sqrt{\nu \pi} \Gamma\left(\frac{\nu}{2}\right)} \left(1+\frac{{d}^{2}}{\nu}\right)^{-\frac{(\nu+1)}{2}},
    \label{eq:t-distribution}
\end{equation}
where $\nu$ denotes the degree of t-distribution freedom. Notice that the t-distribution approximates the Gaussian distribution when $\nu\rightarrow +\infty$, and approximates the uniform distribution when $\nu\rightarrow 0$. The latent space is usually a (normalized) Euclidean space $(Z,d_Z)$. The similarities of input and latent spaces are written as
\begin{align}
    p_{X}(x_i,x_j) &=\alpha_{t} \sum_{t=1}^{m} A_{ij,t} \kappa(d_{X,t}(x_i,x_j),v_{X}), \label{eq:px} \\
    p_{Z}(z_i,z_j) &=\kappa(d_{Z}(z_i,z_j),v_{Z}),
    \label{eq:pz}
\end{align}
where $\alpha_{t}$ is a balancing hyper-parameter for $d_{X,t}$. Notice that $\sum_{i,j} p_{X}(x_i,x_j)$ and $\sum_{i,j} p_{Z}(z_i,z_j)$ are not equal to $1$ in a mini-batch.

\begin{figure}[t!]
    \centering
    \includegraphics[width=0.92\linewidth]{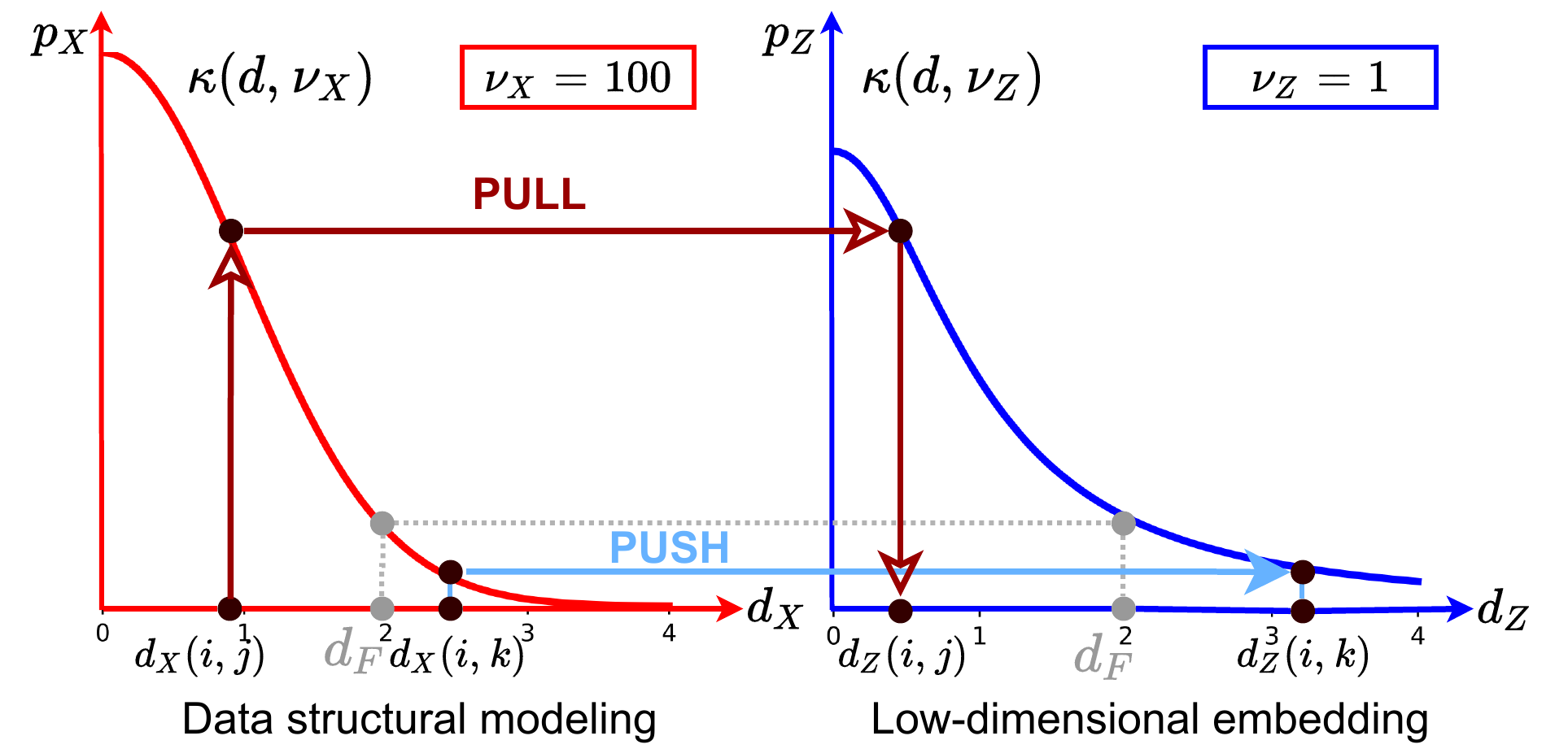}
    \caption{Visualization of $p_X$ and $p_Z$ using the t-distribution.
    Let $d_{Z}^{X} = \kappa^{-1}(\kappa(d_{X}, \nu_{X}), \nu_{Z})$ to be the projected distance of $d_{X}$ to the latent space. When $\nu_X > \nu_Z$, there exists a fix point $d_F$ between the t-distribution with $\nu_X$ and $\nu_Z$, we have $d_{X}(i,j)-d_F < d_{Z}(i,j)-d_F$ (\textit{pull} between neighbors) and $d_{X}(i,k)-d_F < d_{Z}(i,k)-d_F$ (\textit{push} between disjoint samples).}
    \label{fig:t_distribution}
\end{figure}

{\textbf{Data Structual Modeling.}}
Since $p_{X}$ can reflect the reliability of relations between $x$ and other samples, we can control learned representations by the \textit{push} and \textit{pull} forces with various $\nu_{X}$ and $\nu_{Z}$. {Taking the DR task as an example, we assume $d_{X}$ is reliable in the original structure of input data while $d_{Z}$ is likely to be distorted in the extremely low-dimensional space (\textit{e.g.}, 2-dim), as shown in Figure~\ref{fig:t_distribution}: 
We set $\nu_{X}\rightarrow +\infty$ (\textit{i.e.}, the Gaussian distribution) giving the local sample pair $(x_i,x_j)$ and the disjoint sample pair $(x_i,x_k)$ high and low similarities, respectively, while set $\nu_{X}=1$ to make $d_{Z}(x_i,x_j) \ll d_{Z}(x_i,x_k)$.} As explained in Figure~\ref{fig:t_distribution}, the \textit{push} and \textit{pull} forces enable the learned embedding preserving geometric and topological properties of the input data after optimizing Eq.~\ref{eq:final_loss}. 
For practical purposes, we can fix $\nu_{X}$ and adjust $\nu_{Z}$ in $[100, 0)$ to control the structure of the latent space based on the characters of URL tasks (detailed in Sec.~\ref{ch5.5:Summary}).

We provide \textit{static} and \textit{dynamic} methods to adaptively model the similarity $p_{X}$ based on Eq.~\ref{eq:px}. As for the \textit{static}, we first normalize $d_{X,t}$ as $\tilde d_{t}(x_i,x_j) = \frac{d_{t}(x_i,x_j)-\mu_{i,t}}{\sigma_{t}}$, where $\mu_{i,t}$ measures the distance scale of each $x_i$ and $\sigma_{t}$ controls the extension of local neighborhood systems. 
We select proper $\mu_{i,t}$ and $\sigma_{t}$ in terms of population statistics of data (detailed in Sec.~\ref{sec:instantiation}), such as mean and standard deviation of all samples. We rewrite Eq.~\ref{eq:px} and Eq.~\ref{eq:pz} for the \textit{static} $\tilde p_{X}$ as
\begin{align}
    \tilde p_{X}(x_i,x_j) &=\sum_{k=1}^{K}\alpha_{k} \kappa(\frac{d_{k}(x_i,x_j)-\mu_{i,k}}{\sigma_{k}},v_{X}), \label{eq:px_hat} \\
    \tilde p_{Z}(z_i,z_j) &=\kappa(\frac{d_{Z}(z_i,z_j)-\mu_{Z}}{\sigma_{Z}},v_{Z}).
    \label{eq:pz_hat}
\end{align}
Then, we embed $X$ into the latent space $Z$ by minimizing the dis-similarity between $\tilde p_{X}$ and $\tilde p_{Z}$ by a loss function $\mathcal{L}(.)$,
\begin{equation}
    \min_{\theta} \mathcal{L}(\tilde p_{X},\tilde p_{Z}).
    \label{eq:final_loss}
\end{equation}
{\textbf{Low-dimensional Transformation.}}
Notice that the \textit{static} $\tilde p_{X}$ mainly relies on the balancing weight $\alpha_t$ for each $d_{X,t}$, resulting in sub-optimal performances when some distances are unreliable.
{Thus, we design the \textit{dynamic} method utilized in the early stage of the encoder $f_{\theta}$ to achieve reliable low-dimensional transformation, say, the $l$-th layer where $l\in [1,L-1]$.} Since the encoder $f_{\theta}$ will gradually capture data structures by optimizing Eq.\ref{eq:final_loss}, we regard the latent space of the $l$-th stage $\tilde p_{Z,l}$ as the $dynamic$ $\tilde p_{X}$, which adaptively combines various $d_{X,t}$. Based on the \textit{static} $\tilde p_{X}$, we define the \textit{dynamic} $\tilde p'_{X} = \beta\tilde p_{X} + \tilde p_{Z,l}$, where $\beta$ is a weight which linearly decays from $1$ to $0$. Notice that $\tilde p_{Z,l}$ does not require gradient. Finally, our proposed GenURL is demonstrated in Figure~\ref{fig:pipeline}. As discussed in Sec.~\ref{sec:instantiation} and Sec.~\ref{ch5:experiments}, the \textit{static} $\tilde p_{X}$ usually suits URL tasks with well-defined input spaces like DR and KD, while the \textit{dynamic} fits other tasks like CL and GE.

\subsection{Loss Function}
\label{subsec:loss}
As we formulate the URL problem as Eq.~\ref{eq:final_loss} where $\tilde p_{X}$ is regarded as the target, we discuss several similarity functions to achieve optimal embedding. Here, we consider $\tilde p_{X}$ in two cases: (i) generated by incomplete metric spaces where the relationship between distant neighbors is unknown; (ii) generated by well-defined metric spaces. In case (i), we focus on preserving the structures of each neighborhood system (similar pairs). In case (ii), we pay equal attention to dis-similar pairs to capture global relationships. We analyze these losses in various tasks by ablation studies in Sec.~\ref{ch5:experiments}.

\paragraph{Mean squared error (MSE)}
Mean squared error is the most commonly used loss function to measure the similarity between $\tilde p_{X}$ and $\tilde p_{Z}$ with $L_2$-norm,
\begin{equation}
    \mathcal{L}_{MSE}(\tilde p_{X},\tilde p_{Z})=\sum_{i,j} \vert| \tilde p_{X}(x_i,x_j)-\tilde p_{Z}(z_i,z_j)\vert|_{2}^2,
    \label{eq:MSE}
\end{equation}
However, the MSE treats all sample pairs equally, resulting in sub-optimal solutions in both cases.

\paragraph{Kullback-Leibler divergence (KL)}
The KL divergence is commonly used to measure the similarity between two probability distributions,
\begin{equation}
    \mathcal{L}_{KL}(\tilde p_{X},\tilde p_{Z})=-\sum_{i,j} \tilde p_{X}(x_i,x_j) \log \frac{\tilde p_{X}(x_i,x_j)}{\tilde p_{Z}(z_i,z_j)}.
    \label{eq:KL}
\end{equation}
When $\log\tilde p_{X}$ is constant, the KL divergence is equal to the cross-entropy between $\tilde p_{X}$ and $\tilde p_{Z}$. We can regard $\tilde p_{X}$ as a re-weight factor from similar sample pairs to dis-similar sample pairs. However, the KL divergence requires $\tilde p_{Z}(z_i,z_j)=1$~\cite{jmlr2008tSNE}. When $\tilde p_{Z}$ is not a probability distribution~\cite{2018UMAP, eccv2022dlme}, it might result in a trivial solution, $\tilde p_{Z}(z_i,z_j)\rightarrow0$ for each $\tilde p_{X}(x_i,x_j)$.

\paragraph{Binary cross-entropy (BCE)}
To make up the defect in the KL divergence, the binary cross-entropy loss~\cite{2018UMAP} adds a symmetric term for $\tilde p_{X}(x_i,x_j)\rightarrow 0$ (placing higher weights than the KL divergence) to prevent the trivial solution,
\begin{equation}
    \begin{split}
        \mathcal{L}_{BCE}(\tilde p_{X}, \tilde p_{Z})=-\sum_{i,j} \tilde p_{X}(x_i,x_j) \log \tilde p_{Z}(z_i,z_j) -\\
        \sum_{i,j}(1-\tilde p_{X}(x_i,x_j)) \log (1-\tilde p_{Z}(z_i,z_j)).
    \end{split}
    \label{eq:BCE}
\end{equation}
The binary cross-entropy loss optimizes the most similar and dis-similar pairs symmetrically, which is suitable to perverse a well-defined metric space in the case (ii).

\paragraph{General Kullback-Leibler divergence (GKL)}
Although the BCE loss can solve the case (i) well under ideal conditions, it might suffer from outliers, \textit{e.g.,} false negative samples in SSL and GE tasks~\cite{2021iclrHCL, 2022aaaiunmix, Zang2021DMAGE}, resulting in performance degradation in the case (ii). Since we can regard the symmetric term in Eq.~\ref{eq:BCE} as the normalization constrain, $\sum_{i,j}\tilde p_{X}(x_i,x_j)=\sum_{i,j}\tilde p_{Z}(z_i,z_j)$, it is a direct way to prevent the trivial solution in KL divergence. However, the symmetric term emphasizes the importance of the negative samples (dissimilar pairs) with the reweight factor $1-\tilde p_{X}$. Therefore, we propose a relaxed version of the BCE loss with a relaxed symmetric term,
\begin{equation}
    \begin{split}
        \mathcal{L}_{GKL}(\tilde p_{X}, \tilde p_{Z})=-\sum_{i,j} \tilde p_{X}(x_i,x_j) \log\tilde p_{Z}(z_i,z_j) + \\
        \gamma \sum_{i,j}\vert|\tilde p_{X}(x_i,x_j)-\tilde p_{Z}(z_i,z_j)\vert|_{p},
    \end{split}
    \label{eq:GKL}
\end{equation}
where $\gamma$ is a balancing weight and $\vert|.\vert|_{p}$ is $L_p$-norm. We adopt $L_1$-norm and set $\gamma=0.1$ in the GKL loss. Compared to the BCE, the GKL loss is less affected by unreliable samples (usually dis-similar samples) when some $d_{X,t}$ are not reliable.

\section{Instantiation of GenURL}
\label{sec:instantiation}
GenURL generalizes different tasks by fully utilizing partially available information within corresponding datasets and patching the missing properties of URL modeling.

\paragraph{Dimension Reduction and Graph Embedding}
The goal of GE tasks is to encode the geometric and topological structures of the input data. Given a graph $\mathcal{G}=(V,E,X)$, an adjacency matrix $A_1$ can be defined based on $\mathcal{G}$ and another $A_2$ based on a kNN graph built with the feature space $X$. We caculate the shortest-path distance for the entire graph,  $d_{1}(v_i,v_j) = \vert|v_i-v_j\vert|_{2}$ when $A_{1,ij}=1$, and set $d_{1}(v_i,v_j)$ to a large constant when $A_{1,ij}=0$. The distance $d_{2}$ of the kNN graph is obtained in the same way. 
Intuitively, we define the input similarity $p_{X} \triangleq \alpha_{1}p_{1} + \alpha_{2} p_{2}$, where $\alpha_{1}=1$. To remove the scale effects of distances in different spaces, 
we calculate $\mu_{i,t} = \min(d_{t}(x_i,x_0),...,d_{t}(x_i,x_n))$ and $\sigma_{t} = \frac{1}{n}\sum_{i=1}^{n}\sigma_{i}$ in data $X$. Instead of GCN, used in most GE methods, we use $5$-layer MLP using leaky ReLU activation, 
with the middle latent dimension of $512$ and the embedding dimension of $128$. As for the DR task, it is regarded as a special case of GE tasks, which only requires a kNN graph. We adopt the \textit{static} $\tilde p_{X}$ for both the tasks. Similarly, $3$-layer MLP is adopted for DR tasks with the $2$-dim embedding space.

\paragraph{Self-supervised Learning for visual representation}
Unlike the DR and GE tasks, the distance on raw images in the open scenes is unreliable to reflect the desired low-dimensional structures for most discriminative downstream tasks since most images are highly redundant and unstructured. Hence, we import the proxy knowledge of the instance discriminative task in CL~\cite{2020simclr,cvpr2020moco} as follows. Given a mini-batch of data $X^{B}=\{x_i\}_{i=1}^B$, we apply augmentation $\tau\sim\mathcal{T}$ to each sample as $\tau(x_i)$ to obtain two correlated views $X_{a}^{B}=\{x_{i}^{a}\}_{i=1}^B$ and $X_{b}^{B}=\{x_{j}^{b}\}_{j=1}^B$, and fed to the encoder $f_{\theta}$ (\textit{e.g.}, ResNet~\cite{cvpr2016resnet}) and a projection MLP neck~\cite{2020simclr}, denoted as $h_{\phi}$, producing batches of latent representations $Z^{B}_{a},Z^{B}_{b}$ and $H^{B}_{a}$,$H^{B}_{b}$, where $z_i=f_{\theta}(x_i)$ and $h_i=h_{\phi}(z_i)$. The projection neck will be discarded after pre-training. We can convert the proxy knowledge of content invariance into an adjacency matrix $A$: $A_{ii}=1$ for two different views $x_{i}^{1}$ and $x_{i}^{2}$ (positive pairs) of the image $x_{i}$, while $A_{ij}=0$ for any negative image pair $x_i$ and $x_j$. We adopt the $L_{2}$-normalized cosine distance of the projection as the latent representation, $d_{Z} \triangleq \frac{h_i}{\vert|h_i\vert|_2}\cdot \frac{h_j}{\vert|h_j\vert|_2}$. As we discussed in Sec.~\ref{subsec:framework}, we adopt two versions of the input similarity. As for the \textit{static} $\tilde p_{X}$, we use the discrete distance $d_{1}$ defined by the proxy knowledge and the Euclidean distance $d_{2}$ defined by kNN graph in $X$, $p_{X} \triangleq \alpha_{1}p_{1} + \alpha_{2} p_{2}$, where $\alpha_{1}=1$ and $\alpha_{2}=0.01$ which is linearly decreased to 0. 
As for the \textit{dynamic}, we calculate the cosine distance $\frac{z_i}{\vert|z_i\vert|_2}\cdot \frac{z_j}{\vert|z_j\vert|_2}$ and define $\tilde p_{Z,L}$ on the first latent space. The \textit{dynamic} $\tilde p'_{X} = \beta\tilde p_{X} + \tilde p_{Z,L}$.

\paragraph{Unsupervised Knowledge Distillation}
As for the KD task, we regard it as a special type of DR task, \textit{i.e.}, encode the compact latent space of pertained teachers $z^{T}$ into the lower latent space of a student. Since the input space is already a well-defined Euclidean metric space, where the distance of every sample pair can be measured by $d_{Z^T}(x_i,x_j)$, \textit{i.e.}, the input distance $d_{X}\triangleq d_{Z^{T}}$, we adopt the \textit{static} method and use $L_{2}$-normalized cosine distance for the latent space of both the teacher and student. 
To fully explore the knowledge in teacher models, we should pay more attention to the global structural relation between distant samples while preserving the local geometry. As discussed in Sec.~\ref{subsec:loss}, the BCE loss is more suitable for KD tasks.

\section{Experiments}
\label{ch5:experiments}
In this section, we evaluate the effectiveness of GenURL on various unsupervised learning tasks, including self-supervised visual representation (SSL), unsupervised knowledge distillation (KD), graph embedding (GE), and dimension reduction (DR). Meanwhile, we conduct ablation studies of loss functions and hyper-parameters to explore characters of various scenarios.

\subsection{Experimental Setup.}
\label{ch5:exp_setting}
As for evaluation protocols, we adopt the linear protocol as the standard practice~\cite{iccv2019scaling,cvpr2020moco}, which trains a linear classifier on top of fixed representations. As for self-supervised visual representation, we further follow the semi-supervised classification~\cite{2020simclr} and evaluate the generation ability of representations by transfer learning~\cite{cvpr2020moco}. As for dimension reduction, we further adopt trustworthiness (Trust) and continuity (Cont)~\cite{icml2020TopoAE} to measure the distortion between the input data and representations. We use the following training settings for different tasks unless specified. We use Adam optimizer~\cite{iclr2015adam} with a learning rate of $lr\times BatchSize/256$ (linear scaling~\cite{2017msgd}) and a base $lr=0.0005$. The batch size is $256$ by default. All experiments report the mean of $3$ times as default. The best and second results are denoted by \textbf{bold} and \underline{underlined}.

\paragraph{Datasets}
Various types of datasets are used in diverse URL tasks. 
Image datasets include: (1) MNIST~\cite{1998MNIST} contains gray-scale images of 10 classes in 28$\times$28 resolutions, 50K for training, and 10K for testing; (2) FashionMNIST~\cite{2017FMNIST} contains images of 10 classes of fashion clothing (same setting as MNIST); (3) COIL-20~\cite{1996cOIL20} contains 72 different views (over an interval of 3 degrees) for 20 objects, for a total of 1440 images in 128$\times$128 resolutions; (4) CIFAR-10/100~\cite{2009cifar} contains 50K training images and 10K test images in 32$\times$32 resolutions, with 10 classes and 100 classes settings; (5) STL-10~\cite{2011stl10} consists of 5K labeled training images for 10 classes and 100K unlabelled training images and a test set of 8K images in 96$\times$96 resolutions; (6) Tiny-ImageNet (Tiny)~\cite{2017tinyimagenet} has 10K training images and 10K validation images of 200 classes in 64$\times$64 resolutions; (7) ImageNet-1K (IN-1K)~\cite{nips2012imagenet} contains 1.28M training image and 50K validation images from 1000 classes in 224$\times$224 resolutions. Datasets (1-3) are used for DR tasks, and (4-7) are used for SSL and KD tasks. Graph datasets for GE tasks include (8) CORA~\cite{2004cora} contains binary word vectors of 7 classes with 2708 nodes, 1433 features, and 5429 edges; (9) CiteSeer~\cite{1998citeseer} has binary word vectors of 6 classes with 3327 nodes, 3703 features, and 4732 edges; (10) PubMed~\cite{2016PubMed} are associated with tf-idf weighted word vectors with 19717 nodes, 500 features, and 44338 edges for 3 classes.

\paragraph{Implementation of contrastive learning}
We follow MoCo.v2~\cite{2020mocov2} for contrastive learning (CL) pre-training, which adopts ResNet~\cite{cvpr2016resnet} encoder with a two-layer MLP projector based on OpenMixup codebase~\cite{li2022openmixup}. All contrastive learning methods adopt the same network and augmentation settings, while other methods use default settings in their paper. The data augmentation setting in MoCo.v2 is as follows: Geometric augmentations include \textit{RandomResizedCrop} with the scale in $[0.2,1.0]$ and \textit{RandomHorizontalFlip}. Color augmentation include \textit{ColorJitter} with \{brightness, contrast, saturation, hue\} strength of $\{0.4, 0.4, 0.4, 0.1\}$ with an applying probability of $0.8$, and \textit{RandomGrayscale} with an applying probability of $0.2$. Blurring augmentation uses a Gaussian kernel of size $23\times 23$ with a standard deviation uniformly sampled in $[0.1, 2.0]$. As shown in Table~\ref{tab:stl10_ssl} and Table~\ref{tab:cifar_tiny_linear}, we use the GKL loss with $\nu_{X} = \nu_{Z}=100$ and $\sigma_{X}=\sigma_{Z}=0.1$ for GenURL on CIFAR-10, CIFAR-100, STL-10, Tiny ImageNet, and ImageNet-1k datasets.

\paragraph{Implementation of knowledge distillation}
In KD tasks, GenURL follows the settings of the current-proposed contrastive-based KD method SEED~\cite{iclr2021seed}, which adopts the non-linear projector network and data augmentations used in MoCo.v2. Note that MoCo.v2 pre-trained ResNet-50 is adopted as the teacher model. Similar to the SSL task, GenURL uses the BCE loss with $\nu_{X} = \nu_{Z} = 100$ and $\sigma_{X}=1$.
\paragraph{Implementation of graph embedding}
In GE tasks, we adopt $L_{2}$ distance with $\sigma_{Z}=1$ and tune various hyper-parameters as follows. As for $\mu$ and $\sigma$, we perform a grid search of $\mu_{Z}$ and $\sigma_{Z}$ for the latent space in $\{0.01, 0.1, 1, 10, 100\}$ on the validation set. As for $\mu_{i,2}$ and $\sigma_{2}$ of the raw attribute space, we use a binary search (requires $O(n^{2})$) with $5$ nearest neighbors for each data point, \textit{\textit{i.e.}}, the optimal hyper-parameters guarantee the $5$ nearest neighbors of $x_{i}$ have a large similarity score. There are similar practices in UMAP~\cite{2018UMAP} and t-SNE~\cite{jmlr2008tSNE}. If the dataset is too large, we set $\mu_{i,2}$ and $\sigma_{2}$ to the statistic mean and std of the whole dataset.
\paragraph{Implementation of dimension reduction}
GenURL performs DR tasks with the BCE loss and the kNN graph built on the input $X$ following UMAP~\cite{2018UMAP} and DMT~\cite{2021dmt} based on DMT implementation. Similar to the setting of GE tasks, we conduct a grid search of $\nu_{Z}$, $\mu_{Z}$, and $\sigma_{Z}$. We use $\nu_{Z}=0.001$ and $\sigma_{X}=5$ for MNIST and FMNIST datasets while using $\nu_{Z}=0.01$ and $\sigma_{X}=20$ for COIL-20.

\begin{table}[H]
    \vspace{-0.5em}
    \centering
    \caption{\textbf{Linear evaluation and semi-supervised learning on STL-10.} Top-1 accuracy (\%) is reported with various training epochs based on ResNet-50.}
\setlength{\tabcolsep}{1.0mm}
\resizebox{1.0\columnwidth}{!}{
    \begin{tabular}{l|ccc|ccc}
    \toprule
    method                                 & \multicolumn{3}{c|}{Linear}                         & \multicolumn{3}{c}{Semi-supervised}                 \\
                                           & 400ep           & 800ep           & 1600ep          & 400ep           & 800ep           & 1600ep          \\ \hline
    Supervised                             & -               & -               & -               & \gray{71.21}    & \gray{72.70}    & \gray{72.89}    \\
    Related loc~\cite{iccv2015relativeloc} & 60.19           & 64.20           & 64.37           & 86.49           & 87.93           & 88.41           \\
    Rotation~\cite{iclr2018rotation}       & 76.70           & 73.14           & 72.15           & 89.91           & 90.43           & 90.05           \\
    NPID~\cite{eccv2018npid}               & 82.51           & 84.64           & 89.88           & 88.31           & 90.06           & 92.86           \\
    ODC~\cite{cvpr2020odc}                 & 73.43           & 75.47           & 76.20           & 80.88           & 82.04           & 85.80           \\
    SimCLR~\cite{2020simclr}               & \ul{86.92}      & 87.25           & 88.75           & \ul{89.88}      & 90.25           & 91.30           \\
    MoCo.v2~\cite{2020mocov2}              & 84.89           & \ul{89.68}      & \ul{ 91.78}     & 89.66           & \ul{92.53}      & \ul{\bf{93.65}} \\
    BYOL~\cite{nips2020byol}               & 81.17           & 88.74           & 91.41           & 85.38           & 91.71           & 92.69           \\
    SwAV*~\cite{nips2020swav}              & 84.35           & 88.79           & 91.02           & 86.57           & 92.05           & 92.63           \\
    BarlowTwins~\cite{icml2021barlow}      & 85.74           & 88.90           & 91.23           & 86.35           & 91.82           & 92.78           \\
    \rowcolor{gray94}\bf{GenURL}           & \ul{\bf{88.35}} & \ul{\bf{90.82}} & \ul{\bf{91.85}} & \ul{\bf{90.88}} & \ul{\bf{92.58}} & \ul{93.55}      \\
    \bottomrule
    \end{tabular}}
    \label{tab:stl10_ssl}
    \vspace{-0.5em}
\end{table}

\begin{table}[H]
    \vspace{-1.0em}
    \centering
    \caption{\textbf{Linear evaluation on CIFAR-10/100, Tiny ImageNet, and ImageNet-1K.} Top-1 accuracy (\%) are reported. ResNet-18 (R-18) is used as the encoder for CIFAR-10/100 and Tiny ImageNet training 400 and 800 epochs. R-18 and ResNet-50 (R-50) are used for ImageNet-1K training 200 epochs.}
\setlength{\tabcolsep}{0.8mm}
\resizebox{1.0\columnwidth}{!}{
    \begin{tabular}{l|cc|cc|cc|cc}
    \toprule
    method                            & \multicolumn{2}{c|}{CIFAR-10}     & \multicolumn{2}{c|}{CIFAR-100}    & \multicolumn{2}{c|}{Tiny ImageNet} & \multicolumn{2}{c}{ImageNet-1K}   \\
                                      & 400ep           & 800ep           & 400ep           & 800ep           & 400ep            & 800ep           & R-18            & R-50            \\ \hline
    Supervised                        & \gray{94.55}    & \gray{94.89}    & \gray{78.07}    & \gray{78.09}    & \gray{50.68}     & \gray{51.01}    & \gray{69.87}    & \gray{76.56}    \\
    Rotation~\cite{iclr2018rotation}  & 74.81           & 76.32           & 45.52           & 47.82           & 23.46            & 25.17           & 39.85           & 48.25           \\
    NPID~\cite{eccv2018npid}          & 79.53           & 82.70           & 54.52           & 57.16           & 36.86            & 38.24           & 43.10           & 58.87           \\
    ODC~\cite{cvpr2020odc}            & 78.23           & 79.91           & 48.04           & 52.17           & 27.30            & 28.79           & 45.17           & 53.40           \\
    SimCLR~\cite{2020simclr}          & \ul{86.22}      & 88.24           & 56.45           & 57.42           & \ul{37.64}       & 38.46           & 51.03           & 66.67           \\
    MoCo.v2~\cite{2020mocov2}         & 82.41           & \ul{88.62}      & 56.65           & 61.48           & 33.00            & 37.49           & 52.87           & 67.85           \\
    BYOL~\cite{nips2020byol}          & 82.61           & 88.15           & \ul{58.32}      & \ul{\bf{64.40}} & 33.93            & \ul{\bf{38.81}} & \ul{54.62}      & \ul{71.88} \\
    BarlowTwins~\cite{icml2021barlow} & 82.28           & 88.36           & 56.72           & 61.92           & 33.27            & 38.34           & 53.23           & 71.66           \\
    \rowcolor{gray94}\bf{GenURL}      & \ul{\bf{88.27}} & \ul{\bf{88.95}} & \ul{\bf{59.01}} & \ul{63.51}      & \ul{\bf{37.81}}  & \ul{38.68}      & \ul{\bf{55.12}} & \ul{\bf{72.15}}  \\
    \bottomrule
    \end{tabular}
    }
    \label{tab:cifar_tiny_linear}
    \vspace{-0.5em}
\end{table}

\subsection{Self-supervised Visual Representation}
\label{ch5.1:ssl}
In this subsection, we compare GenURL with three types of existing self-supervised methods, including head-craft, clustering-based, and contrastive learning methods. For a fair comparison, we apply the same augmentation settings described in MoCo.v2~\cite{2020mocov2} to all contrastive learning methods and follow hyper-parameters described in their original papers. We remove the Gaussian blur augmentation in CIFAR experiments~\cite{nips2020byol, cvpr2021simsiam}. We perform unsupervised pre-training using ResNet-50 on STL-10 and ImageNet-1K, and using ResNet-18 on CIFAR-10/100, Tiny ImageNet, and ImageNet-1K. GenURL adopts the GKL loss and the \textit{dynamic} method in the SSL task.

\paragraph{Evaluation protocols}
As for linear evaluation, we follow the experiment settings in \cite{cvpr2020moco} to train a linear classifier for 100 epochs and use different base $lr$ for different datasets. We set the base $lr=1.0$ for STL-10 and CIFAR-100, $lr=0.1$ for CIFAR-10 and Tiny, and $lr=30$ for IN-1K. The learning rate decays by $0.1$ at $60$ and $80$ epochs. As for semi-supervised evaluation, we fine-tune the whole pre-trained model for $20$ epochs on STL-10 with a step schedule at $12$ and $16$ epochs, and batch size is $256$. We perform grid search for each test methods on base $lr=\{0.1,0.01,0.001\}$ and parameter-wise $lr\_mul=\{1,10,100\}$ of the $fc$ layer. Both experiments use the SGD optimizer with the weight decay of $0$ for linear evaluation and $0.0001$ for semi-supervised. Top-1 and top-5 accuracy are reported on the validation set.

\paragraph{Linear evaluation results}
We first compare with existing methods in terms of different training epochs on STL-10, as shown in Table \ref{tab:stl10_ssl}. The proposed GenURL achieves the highest accuracy among all settings. It not only converges faster than other algorithms under 400-epoch pre-training but gains better performance when training longer. Then, we compare various methods on CIFAR-10/100 and Tiny, as shown in Table \ref{tab:cifar_tiny_linear}. GenURL achieves the top performance on three datasets under 400-epoch pre-training and achieves second-best on CIFAR-100 and Tiny-ImageNet when training 800 epochs. Different from the existing contrastive-based methods, GenURL takes more pair-wise relations between samples into consideration, which might help GenURL convergence faster. For example, given a mini-batch of $N$ samples, BYOL utilizes $2N$ sample pairs, and MoCo requires $K+N$ sample pairs ($K$ is the memory bank size), while GenURL optimizes $N^2$ sample pairs (similar to SimCLR~\cite{2020simclr}).

\paragraph{Semi-supervised evaluation results}
In Table~\ref{tab:stl10_ssl}, we fine-tune a ResNet-50 pre-trained with various methods on the labeled training set of STL-10. GenURL outperforms other methods under 400-epoch and 800-epoch pre-training, which reflects its fast convergence speeds while maintaining the second-best classification accuracy with longer training.

\paragraph{Transferring Features}
The main goal of unsupervised learning is to learn transferrable features. In Table~\ref{tab:stl10_transfer}, we compare the representation quality of unsupervised pre-training on STL-10 by transferring to the classification task. We adopt linear evaluation on the CIFAR-10 in 64$\times$64 resolutions with 1600-epoch pre-trained ResNet-50 on STL-10, and other settings are the same as Sec.~\ref{ch5.1:ssl}. GenURL achieves the highest accuracy among all methods: +3.36\%/+3.29\%/+2.99\% for GenURL pre-training 400/800/1600 epochs over the second-best method.

\begin{figure}[htb]
    \vspace{-1.0em}
    \centering
    \includegraphics[width=1.0\linewidth]{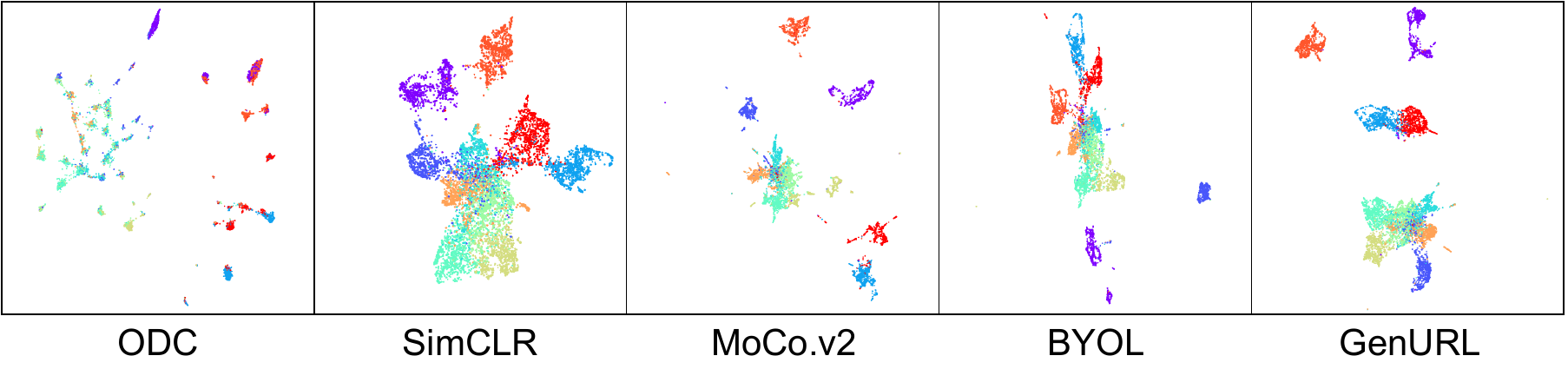}
    \vspace{-2.0em}
    \caption{Visualization of learned representations of CL methods with ResNet-50 on STL-10. We visualize the 2048-dim embeddings by UMAP. Compared to ODC and BYOL, the local structures of clusters are well-preserved, while each cluster is discriminative.}
    \label{fig:ssl_vis}
    \vspace{-1.0em}
\end{figure}

\paragraph{Ablation studies for SSL tasks}
We first ablate the loss functions used in visual SSL tasks. Since the input distance in SSL tasks is only well-defined for positive pairs where it can be optimized explicitly, and the relationship between negative pairs can be implicitly modeled by the \textit{dynamic} method. As shown in Table~\ref{tab:ssl_loss_ablation}, GenURL prefers the GKL loss when using the \textit{static} $\tilde p_{X}$, while the BCE loss yields better performance when using the \textit{dynamic} to mine the relation between negative pairs. 
Then, we analyze hyperparameters of GenURL in Figure \red{5}.
We find that GenURL prefers $\nu_{Z}=100$ and $\sigma=1$ (approximating a standard Gaussian kernel) with a small batch size of 256. Moreover, we compare learned representations of GenURL with other visual SSL methods on STL-10 by UMAP~\cite{2018UMAP} visualization in Figure~\ref{fig:ssl_vis}.

\begin{table}[t]
    \centering
    \vspace{-0.25em}
\caption{\textbf{Loss function analysis on self-supervised learning.} We evaluate the loss functions proposed in Sec. 3 on STL-10, CIFAR-10/100, and Tiny ImageNet. Top-1 accuracy (\%) under linear evaluation is reported.}
\resizebox{0.93\columnwidth}{!}{
    \begin{tabular}{llcccc}
    \toprule
    Loss  & $\widetilde p_{X}$ setting & STL-10         & CIFAR-10       & CIFAR-100      & Tiny           \\ \hline
    MSE   & \textit{static}        & 88.72          & 84.26          & 57.31          & 36.82          \\
    BCE   & \textit{static}        & 91.05          & 88.35          & 60.08          & 38.19          \\
    BCE   & \textit{dynamic}       & 91.60          & 88.87          & 61.16          & \textbf{38.85} \\
    GKL   & \textit{static}        & 91.21          & 88.63          & 61.27          & 38.07          \\
    GKL   & \textit{dynamic}       & \textbf{91.85} & \textbf{88.95} & \textbf{61.51} & 38.48          \\
    \bottomrule
    \end{tabular}}
    \label{tab:ssl_loss_ablation}
    \vspace{-1.0em}
\end{table}

\begin{figure}[htb]
    \centering
    \includegraphics[width=1.0\linewidth]{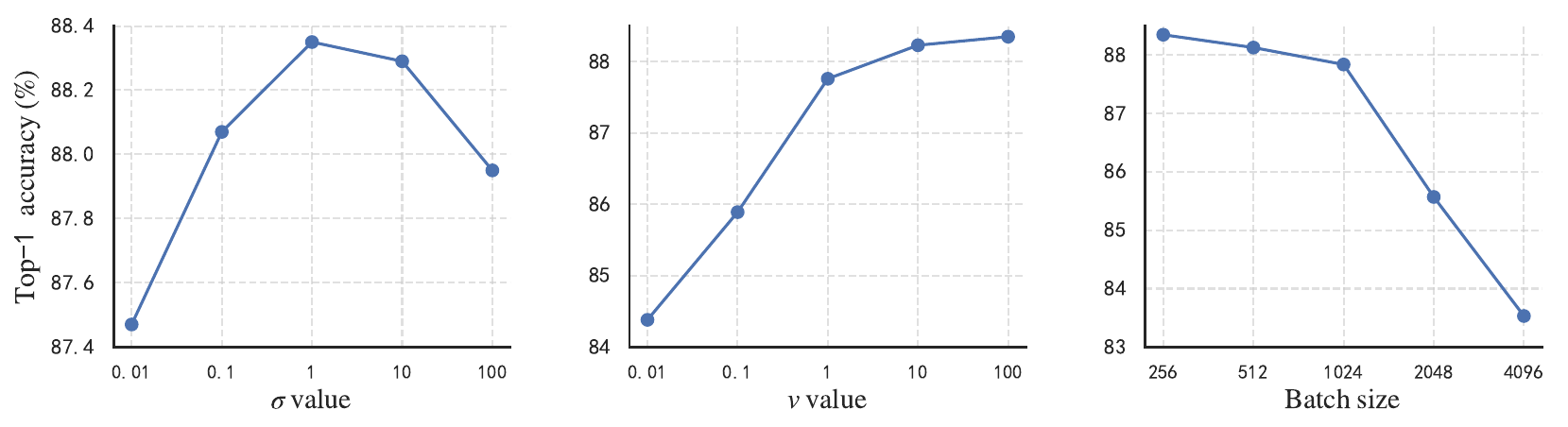}
    \label{fig:ssl_ablation}
    \vspace{-2.25em}
    \caption{Ablation of $\nu_{Z}$, $\sigma$ and batch size of GenURL for visual SSL tasks on STL-10. GenURL is pre-trained 800-epoch with ResNet-50.}
\end{figure}

\begin{figure*}
\begin{minipage}{0.695\linewidth}
\centering
\begin{table}[H]
\centering
\caption{\textbf{Unsupervised knowledge distillation.} Top-1 accuracy (\%) under linear evaluation on STL-10. The teacher model is ResNet-50 pre-trained by MoCo.v2. $\dag$ indicates using a momentum encoder as MoCo.v2. SSL denotes the InfoNCE loss. KD denotes the knowledge distillation loss. H+AW denotes the Huber loss and angle-preserving loss in RKD.}
\setlength{\tabcolsep}{1.2mm}
\resizebox{1.0\columnwidth}{!}{
\begin{tabular}{ll|ccc|ccc|ccc}
    \toprule
KD                                  & KD         & \multicolumn{3}{c|}{ResNet-18}            & \multicolumn{3}{c|}{MobileNet.v2}         & \multicolumn{3}{c}{ShuffleNet.v1}         \\
methods                             & loss       & SSL   & KD              & KD+SSL          & SSL   & KD              & KD+SSL          & SSL   & KD              & KD+SSL          \\ \hline
RKD\cite{cvpr2019rkd}               & H+AW       &       & 86.48           & 86.76           &       & 85.89           & 86.20           &       & 84.31           & 85.01           \\
PKT\cite{eccv2018pkt}               & KL         &       & 86.89           & 87.12           &       & 86.14           & 86.48           &       & 84.25           & 84.82           \\
SP\cite{iccv2019sp}                 & MSE        &       & 86.53           & 86.74           &       & 85.96           & 86.13           &       & 84.22           & 84.76           \\
SSKD\cite{eccv2020SSKD}             & KL+InfoNCE & 81.51 & -               & 87.78           & 79.96 & -               & 86.80           & 77.26 & -               & 85.23           \\
CRD\cite{iclr2020CRD}               & KL+InfoNCE &       & -               & 87.24           &       & -               & 86.39           &       & -               & 84.98           \\
SEED$^\dag$~\cite{iclr2021seed}     & KL+InfoNCE &       & -               & 87.36           &       & -               & 86.44           &       & -               & 85.02           \\
\rowcolor{gray94}\bf{GenURL}        & BCE        &       & \ul{88.05}      & \ul{88.13}      &       & \ul{86.61}      & \ul{86.85}      &       & \ul{84.67}      & \ul{85.10}      \\
\rowcolor{gray94}\bf{GenURL}$^\dag$ & BCE        &       & \ul{\bf{88.26}} & \ul{\bf{88.39}} &       & \ul{\bf{87.28}} & \ul{\bf{87.47}} &       & \ul{\bf{85.05}} & \ul{\bf{85.38}} \\
    \bottomrule
    \end{tabular}
    }
    \label{tab:kd_stl10_linear}
\end{table}

\end{minipage}
\begin{minipage}{0.30\linewidth}
\centering
\begin{table}[H]
\centering
\caption{\textbf{Transfer learning on CIFAR-10 classification}. Top-1 accuracy (\%) under linear evaluation is reported.}
\setlength{\tabcolsep}{1.0mm}
\resizebox{1.0\columnwidth}{!}{
    \begin{tabular}{lccc}
    \toprule
    method                                 & 400 ep          & 800 ep          & 1600 ep          \\ \hline
    Related loc~\cite{iccv2015relativeloc} & 66.41           & 69.34           & -                \\
    Rotation~\cite{iclr2018rotation}       & 71.01           & 64.29           & -                \\
    NPID~\cite{eccv2018npid}               & 71.15           & 63.72           & 65.30            \\
    ODC~\cite{cvpr2020odc}                 & 68.59           & 66.13           & 70.51            \\
    SimCLR~\cite{2020simclr}               & \ul{75.97}      & 75.08           & \ul{76.86}       \\
    MoCo.v2~\cite{2020mocov2}              & 74.46           & 76.54           & 75.61            \\
    BYOL~\cite{nips2020byol}               & 74.04           & \ul{76.83}      & 75.55            \\
    SwAV*~\cite{nips2020swav}              & 74.17           & 76.28           & 76.34            \\
    BarlowTwins~\cite{icml2021barlow}      & 74.63           & 76.71           & 76.12            \\
    \rowcolor{gray94}\bf{GenURL}           & \ul{\bf{80.22}} & \ul{\bf{80.12}} & \ul{\bf{79.85}}  \\
    \bottomrule
    \end{tabular}}
    \label{tab:stl10_transfer}
\end{table}

\end{minipage}
\end{figure*}

\subsection{Unsupervised Knowledge Distillation}
\label{ch5.2:KD}
We evaluate the KD tasks based on self-supervised learning on STL-10 dataset. In this experiment, we adopt MoCo.v2 with ResNet-50 under 1600-epoch pre-training. We choose multiple smaller networks with fewer parameters as the student network: ResNet-18~\cite{cvpr2016resnet}, MobileNet.v2~\cite{cvpr2018mobilenetv2}, ShuffleNet.v1~\cite{zhang2017shufflenet}. Similar to the pre-training for the teacher network, we add one additional MLP layer on the basis of the student network. Follow the linear evaluation protocols in Sec.~\ref{ch5.1:ssl}, we compare the existing relation-based KD methods including RKD~\cite{cvpr2019rkd}, PKT~\cite{eccv2018pkt}, SP~\cite{iccv2019sp}, SSKD~\cite{eccv2020SSKD}, CRD~\cite{iclr2020CRD}, and SEED~\cite{iclr2021seed}. We adopt the BCE loss for GenURL in the KD task.

\paragraph{Linear evaluation results}
From the view of different student models, as shown in Table~\ref{tab:kd_stl10_linear}, we notice that smaller networks perform rather worse and also benefit more from distillation than larger networks. From the perspective of various KD loss functions, the results clearly demonstrate that the proposed GenURL with the BCE loss achieves the best results, which is mainly because the BCE loss optimizes both local geometries and global structures.

\begin{figure}[b]
    \centering
    \includegraphics[width=1.0\linewidth]{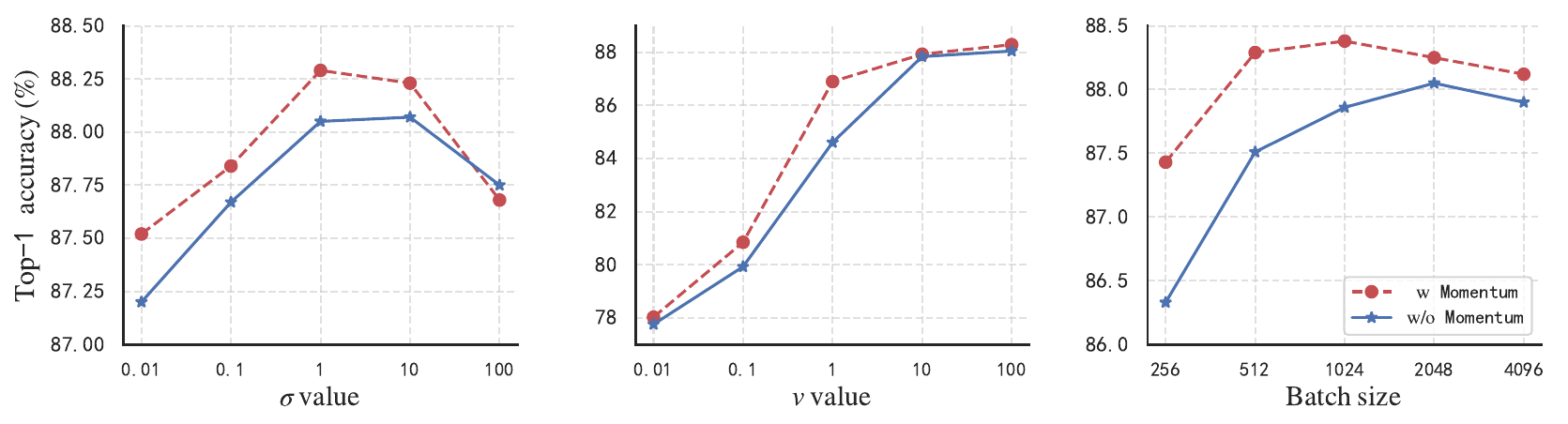}
    \vspace{-2.0em}
    \caption{Ablation of $\nu_{Z}$, $\sigma$ and batch size of GenURL for KD tasks on STL-10.}
    \label{fig:kd_ablation}
\end{figure}

\paragraph{Ablation studies for KD tasks}
In contrast to SSL tasks, the input distance in KD tasks is assumed to be well-defined for both positive and negative sample pairs. Thus, we adopt the \textit{static} $\tilde p_{X}$ and try to mine relationships among different sub-manifolds by momentum memory bank (\textit{M}) for negative samples, which is similar to dark knowledge~\cite{nips2014kd} in supervised KD tasks. 
Then, we ablate hyperparameters of GenURL for KD tasks in Figure~\ref{fig:kd_ablation}. We find that using the BCE loss and \textit{M} with a large batch size achieves the best performance, and the GKL loss with \textit{M} yields the best result when using a small batch size.

\begin{table}[b]
    \vspace{-0.5em}
    \centering
    \vspace{-0.25em}
\caption{\textbf{Loss function analysis on unsupervised knowledge distillation.} We evaluate the loss functions proposed in Sec. 3 on STL-10. Top-1 accuracy (\%) under linear evaluation is reported. M denotes using the momentum encoder.}
\resizebox{0.90\columnwidth}{!}{
    \begin{tabular}{ccccccc}
    \toprule
    batch & \multicolumn{2}{c}{MSE}        & \multicolumn{2}{c}{BCE}        & \multicolumn{2}{c}{GKL}        \\ \cline{2-7} 
    size  & w/o \textit{M} & w/ \textit{M} & w/o \textit{M} & w/ \textit{M} & w/o \textit{M} & w/ \textit{M} \\ \hline
    256   & 85.98          & 86.10         & 84.11          & 87.37         & 86.33          & \bf{88.02}    \\
    2048  & 86.14          & 86.43         & 88.05          & \bf{88.26}    & 85.97          & 85.91         \\
    \bottomrule
    \end{tabular}}
    \label{tab:kd_loss_ablation}
\end{table}

\subsection{Unsupervised Graph Embedding}
\label{ch5.3:GraphEmbedding}
\paragraph{Setups and results}
Unsupervised graph embedding experiments are conducted on three graph network datasets (Cora, CiteSeer, and PubMed), and we evaluate the learned embeddings by the node classification task. We compare GE methods that utilize both features and graph structures, including AGC~\cite{2019AGC}, AGE~\cite{kdd2020AGE}, GIC~\cite{2020GIC}, and ARGA~\cite{2020ARGA}. The learned node embeddings are passed to logistic regression, and we report the mean and std of linear classification accuracy (Acc) of comparison methods. Table~\ref{tab:node_classification} shows that GenURL (BCE) using \textit{both} graphs and attributed features achieves new state-of-the-art performances on three GE datasets and improves previous GE methods by at least 0.5\%, 0.4\%, and 1.0\% top-1 accuracy on CORA, CiteSeer, and PubMed datasets.
\begin{table}[h]
\centering
\caption{\textbf{Node classification.} Top-1 accuracy (\%) under linear evaluation on CORA, CiteSeer, and PubMed. \textit{both} indicates using both the graph structure and attributed features.}
\setlength{\tabcolsep}{1.3mm}
\resizebox{1.0\columnwidth}{!}{
    \begin{tabular}{llccc}
    \toprule
    Method                            & Input             & CORA                        & CiteSeer                    & PubMed                      \\ \hline
    DeepWalk~\cite{kdd2014DeepWalk}   & \textit{graph}    & 43.68$_{\pm0.04}$           & 36.57$_{\pm0.05}$           & 60.29$_{\pm0.03}$           \\
    AGC~\cite{2019AGC}                & \textit{both}     & 75.50$_{\pm0.02}$           & 67.37$_{\pm0.01}$           & 67.70$_{\pm0.02}$           \\
    AGE~\cite{kdd2020AGE}             & \textit{both}     & 77.77$_{\pm0.02}$           & 61.30$_{\pm0.02}$           & 71.15$_{\pm0.0}$            \\
    GIC~\cite{2020GIC}                & \textit{both}     & 77.27$_{\pm0.01}$           & 41.88$_{\pm0.03}$           & 76.70$_{\pm0.02}$           \\
    ARGA~\cite{2020ARGA}              & \textit{both}     & 72.52$_{\pm0.02}$           & 55.21$_{\pm0.01}$           & 64.24$_{\pm0.02}$           \\
    \rowcolor{gray94}\bf{GenURL(BCE)} & \textit{feature}  & 60.33$_{\pm0.03}$           & 43.78$_{\pm0.08}$           & 70.47$_{\pm0.02}$           \\
    \rowcolor{gray94}\bf{GenURL(GKL)} & \textit{both}     & \ul{78.17$_{\pm0.01}$}      & \ul{67.63$_{\pm0.02}$}      & \ul{77.54$_{\pm0.02}$}      \\
    \rowcolor{gray94}\bf{GenURL(BCE)} & \textit{both}     & \ul{\bf{78.32$_{\pm0.02}$}} & \ul{\bf{67.70$_{\pm0.01}$}} & \ul{\bf{77.75$_{\pm0.02}$}} \\
    \bottomrule
    \end{tabular}}
    \label{tab:node_classification}
\end{table}

\begin{figure}[htb]
    \centering
    \includegraphics[width=1.0\linewidth]{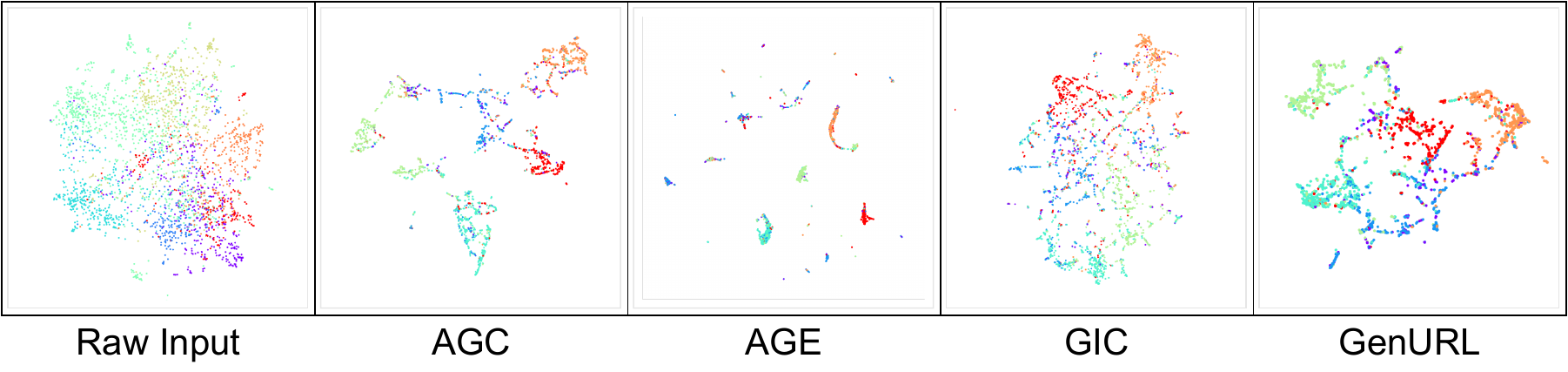}
    \vspace{-2.0em}
    \caption{Visualization of the learned representations of GE methods on Wiki. We visualize the last latent space of the encoder by UMAP. The result of GenURL contains both the topology and the local geometric structures.}
    \label{fig:ge_vis}
\end{figure}

\begin{figure}[htb]
    \centering
    \includegraphics[width=1.0\linewidth]{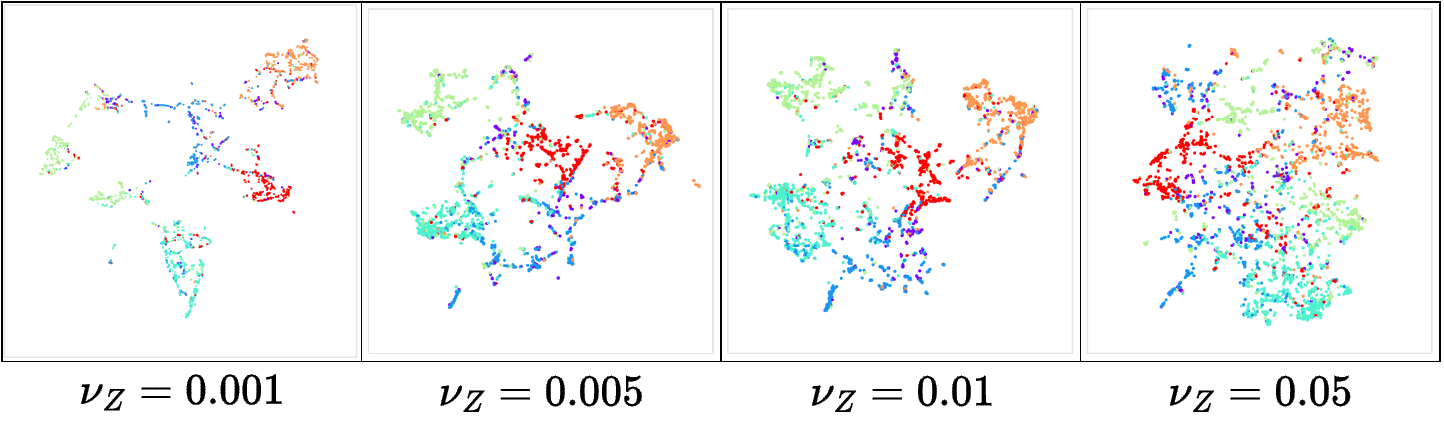}
    \vspace{-2.0em}
    \caption{Ablation of various $\nu_{Z}$ in GenURL for the GE task on CiteSeer. The learned embedding is visualized by UMAP to 2-dim space.}
    \label{fig:ge_ablation}
    \vspace{-0.50em}
\end{figure}

\paragraph{Ablation and analysis}
We first adopt two ablation studies of the loss functions used in GenURL on GE tasks in Table~\ref{tab:node_classification}: (i) when the input is \textit{both} (using both graphs and attributed features with the \textit{dynamic} $\tilde p_{X}$), the BCE loss shows better performance than the GKL loss; (ii) when using the BCE loss, using \textit{both} with the \textit{dynamic} $\tilde p_{X}$ outperforms \textit{feature} (only using the attributed features) with the \textit{static} $\tilde p_{X}$. 
Then, we visualize the learned embedding on CiteSeer by UMAP in Figure~\ref{fig:ge_vis}. We find that GenURL separates sub-graphs of different classes into different clusters while maintaining the local geometric details of each sub-graph. As shown in Figure~\ref{fig:ge_ablation}, we empirically show that $\nu_{Z}$ can control the balance between local geometric and global topological structures, and $\nu_{Z}=0.005$ yields the best visualization results.


\subsection{Dimension Reduction}
\label{ch5.4:DR}
\begin{figure}[ht]
    \centering
    \includegraphics[width=1.0\linewidth]{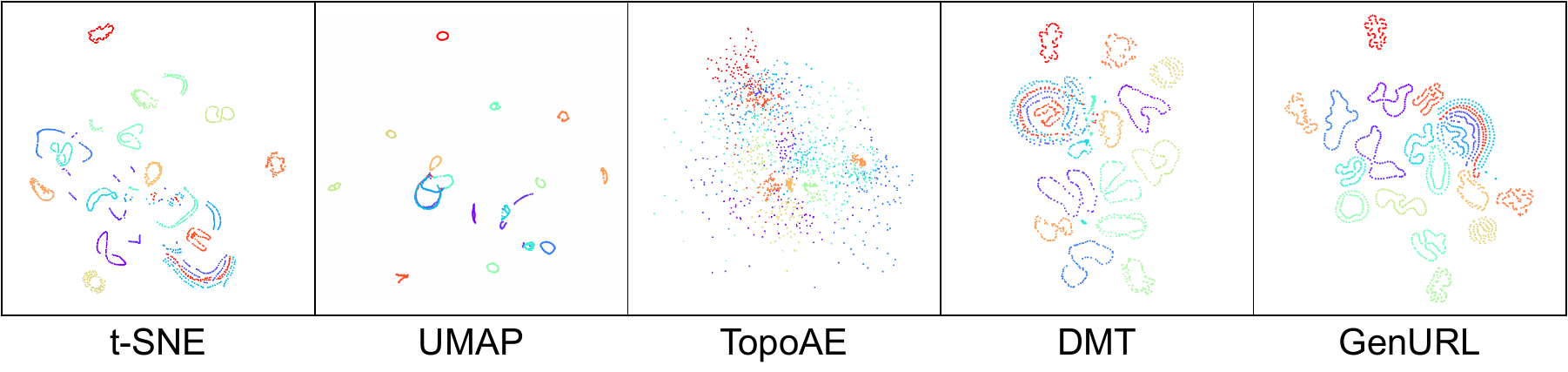}
    \vspace{-2.0em}
    \caption{Visualization of COIL-20 by DR methods to 2-dim space. Compared to UMAP and DMT, GenURL preserves more local geometric details while capturing global topological structures.}
    \label{fig:dr_vis}
\end{figure}

\begin{figure}[ht]
    \centering
    \includegraphics[width=1.0\linewidth]{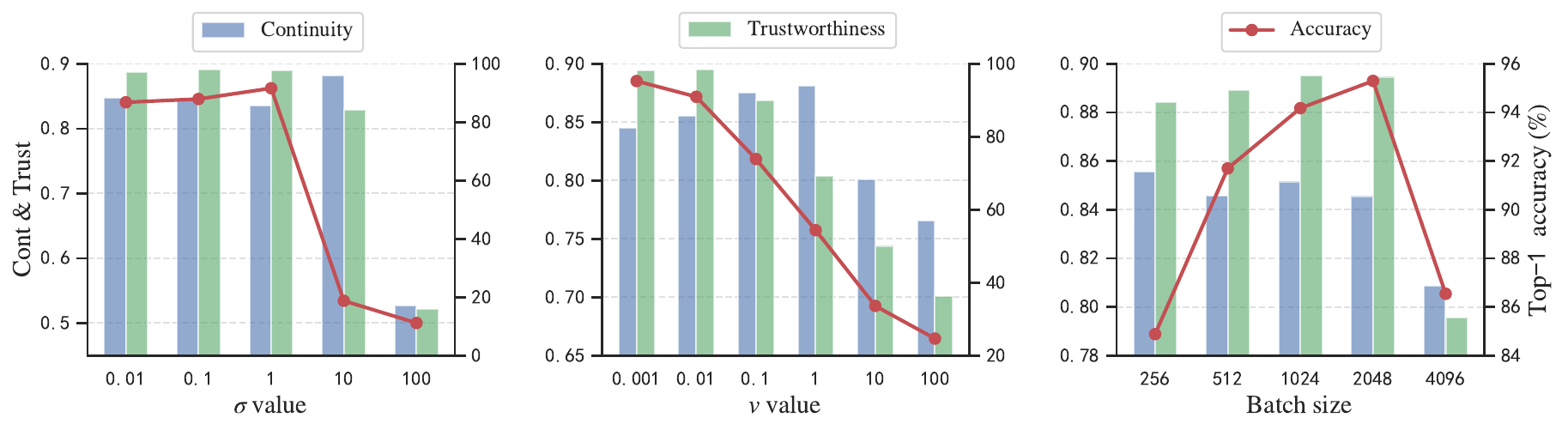}
    \vspace{-2.0em}
    \caption{Ablation of $\nu_{Z}$, $\sigma$ and batch size of GenURL (BCE) for DR tasks on MNIST.}
    \label{fig:dr_hyper_ablation}
\end{figure}

\begin{figure}[tb]
    \centering
    \includegraphics[width=1.0\linewidth]{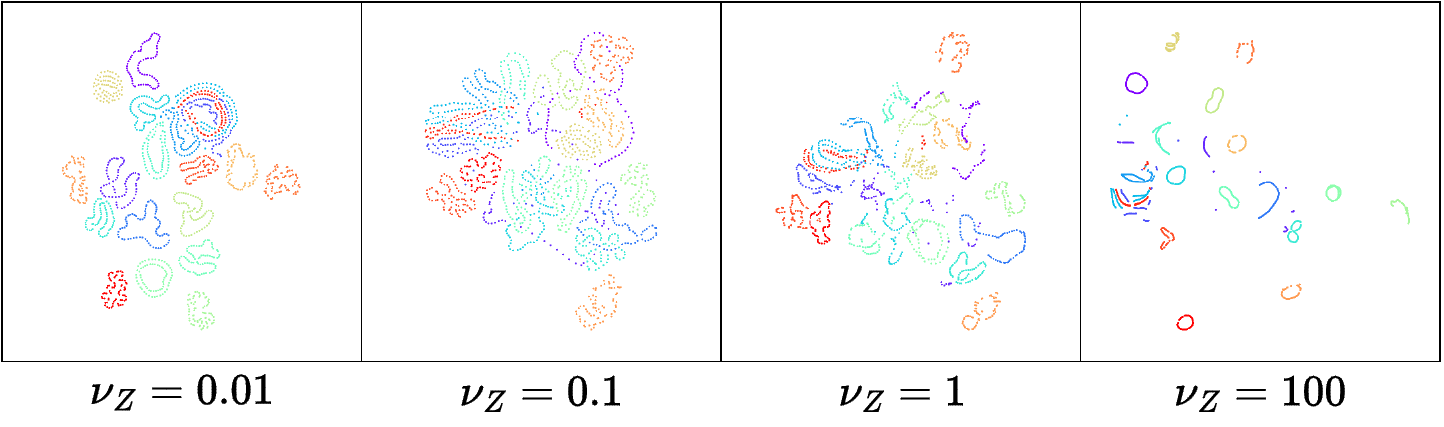}
    \vspace{-1.5em}
    \caption{Ablation of various $\nu_{Z}$ in GenURL for the DR task on CiteSeer. The learned embedding is visualized by UMAP to 2-dim space.}
    \label{fig:dr_ablation}
    \vspace{-1.0em}
\end{figure}

\paragraph{Setups and results}
We perform DR experiments on MNIST, FMNIST, and COIL-20 datasets. We compare the current leading methods, including non-parametric methods (t-SNE~\cite{jmlr2008tSNE} and UMAP~\cite{2018UMAP}) and parametric methods (P-UMAP~\cite{2021pUMAP}, GRAE~\cite{2020GRAE}, TopoAE~\cite{icml2020TopoAE}, and DMT~\cite{2021dmt}). Besides the linear classification top-1 accuracy (Acc) with logistic regression, we evaluate the qualities of the low-dimensional representation in terms of the input space with Trust and Cont. Since the DR task only relies on the input space, GenURL adopts the BCE loss and the \textit{static} $\tilde p_{X}$. As shown in Table~\ref{tab:dr_mnist_fmnist}, we compare GenURL (BCE) with existing DR methods and find that GenURL yields comparable performance in terms of Trust and Acc, which indicates that GenURL (BCE) keeps the balance between local geometric structures (achieving better Trust and Cont) and the distinction of different sub-manifolds (achieving better linear classification accuracy).
\paragraph{Ablation and analysis}
We first conduct the ablation of the BCE or GKL losses in Table~\ref{tab:dr_mnist_fmnist}: GenURL (BCE) always outperforms GenURL (GKL) on three DR datasets because we adopt a large batch size as DMT and P-UMAP. 
Then, we provide DR results on COIL-20 in Figure~\ref{fig:dr_vis} and find that GenURL captures more geometric structures than previous methods, especially UMAP and TopoAE (focusing on topological structures). Moreover, we ablate hyperparameters of GenURL (BCE) on MNIST in Figure~\ref{fig:dr_hyper_ablation} and find that GenURL prefers $\sigma=1$, $\nu_{Z}=0.001$, and the batch size of 2048. Similar to GE tasks, we verify whether $\nu_{Z}$ can control the balance of local structures and global topology by providing visualization of GenURL with various $\nu_{Z}$ on COIL-20. In Figure~\ref{fig:dr_ablation}, we find that $\nu_{Z}=0.01$ yields the best balance between local geometric and global topological structures.

\begin{table}[h]
\centering
\caption{\textbf{Dimension reduction.} Trust, Cont, and top-1 accuracy (\%) are reported on MNIST, FMNIST, and COIL-20.}
\setlength{\tabcolsep}{1.1mm}
\resizebox{1.0\columnwidth}{!}{
    \begin{tabular}{l|ccc|ccc|ccc}
    \toprule
    method                            & \multicolumn{3}{c|}{MNIST}                         & \multicolumn{3}{c|}{FMNIST}                        & \multicolumn{3}{c}{COIL-20}                        \\
                                      & Trust           & Cont            & Acc            & Trust           & Cont            & Acc            & Trust           & Cont            & Acc            \\ \hline
    t-SNE\cite{jmlr2008tSNE}          & 0.872           & 0.855           & 83.4           & 0.949           & 0.934           & 69.2           & \ul{\bf{0.925}} & 0.867           & 89.0           \\
    UMAP\cite{2018UMAP}               & 0.887           & 0.837           & 96.6           & 0.947           & 0.938           & 68.7           & 0.922           & 0.849           & 87.1           \\
    P-UMAP\cite{2021pUMAP}            & 0.890           & 0.834           & \ul{96.7}      & 0.951           & 0.937           & 68.9           & \ul{0.924}      & 0.843           & 89.3           \\
    GRAE\cite{2020GRAE}               & 0.876           & \ul{0.861}      & 75.0           & 0.949           & \ul{0.943}      & 58.2           & 0.920           & 0.887           & 86.5           \\
    TopoAE\cite{icml2020TopoAE}       & 0.881           & \ul{\bf{0.876}} & 75.5           & 0.952           & \ul{\bf{0.970}} & 60.6           & 0.877           & 0.898           & 85.8           \\
    DMT\cite{2021dmt}                 & 0.896           & 0.840           & \ul{96.7}      & \ul{0.958}      & 0.939           & \ul{70.0}      & 0.916           & \ul{0.928}      & 90.2           \\
    \rowcolor{gray94}\bf{GenURL(GKL)} & \ul{\bf{0.897}} & 0.849           & 96.4           & \ul{0.958}      & 0.937           & \ul{70.0}      & 0.923           & 0.926           & \ul{90.3}      \\
    \rowcolor{gray94}\bf{GenURL(BCE)} & \ul{0.898}      & 0.842           & \ul{\bf{96.8}} & \ul{\bf{0.959}} & 0.940           & \ul{\bf{70.2}} & \ul{0.924}      & \bf{\ul{0.930}} & \bf{\ul{90.4}} \\
    \bottomrule
    \end{tabular}}
    \label{tab:dr_mnist_fmnist}
\end{table}


\subsection{Analysis and Discussion}
\label{ch5.5:Summary}
We provide an empirical analysis of the hyper-parameters and loss functions in GenURL on different URL tasks to demonstrate the characteristics of various tasks. We compare the results using different batch sizes, $\nu_{Z}$, $\sigma$, and loss functions used in GenURL to demonstrate the relationship of SSL, KD, and DR tasks.

\paragraph{Relationship between DR and GE}
As shown in Figure~\ref{fig:ge_ablation} and Figure~\ref{fig:dr_ablation}, GenURL prefers smaller $\nu_{Z}$ for both the GE and DR tasks because the large $\nu_{Z}$ yields crowd embedding while the small $\nu_{Z}$ conducts separable results. Therefore, we consider DR as a special type of GE task. The main difference between DR and GE tasks is GE takes both the node and edge features into consideration.

\paragraph{Relationship between SSL and KD}
Then, we compare how GenURL deals with the negative samples in SSL and KD tasks. In Figure~\ref{fig:kd_ablation}, we find that GenURL prefers the similar $\nu_{Z}$ and $\sigma$ for both SSL and KD tasks, which indicates using $\nu_{Z}=100$ and $\sigma=1$ (the standard Gaussian kernel) is suitable for $L_2$-normalized cosine distance. Notice that GenURL prefers small batch sizes like $256$ for the SSL task (suffering performance drops when the batch size increases) while prefers larger batch sizes for the KD task. It might be because pair-wise similarities of negative samples in SSL tasks are unreliable and can be regarded as dark knowledge in the KD task~\cite{nips2014kd, iclr2021seed}. The gradient from negative pairs might overwhelm positive samples at the early training stage of the SSL task, while negative samples are well-defined by the teacher model in the KD task. Meanwhile, Table~\ref{tab:kd_loss_ablation} shows that using the \textit{dynamic} version and the GKL loss in SSL tasks yield the best performance while using the large batch size, and the BCE loss in KD tasks performs better. Therefore, We conclude that the GKL with the \textit{dynamic} structural modeling can alleviate the harmful effects of unreliable metric spaces of SSL tasks.

\begin{figure}[ht]
    \centering
    \includegraphics[width=1.0\linewidth]{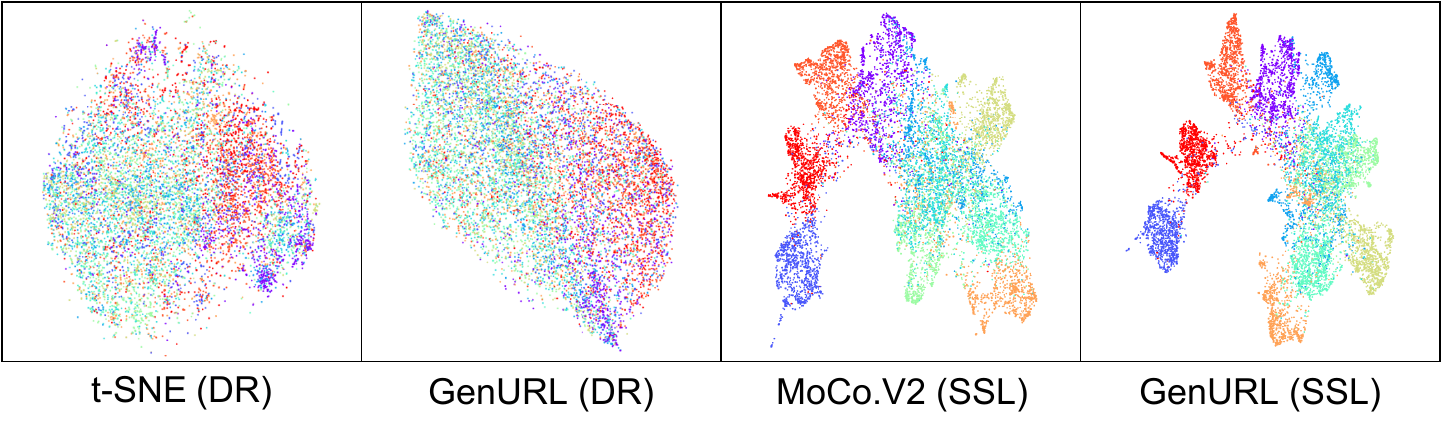}
    \vspace{-2.0em}
    \caption{Visualization of learned representation in DR and SSL tasks on CIFAR-10. The SSL representation is visualized by UMAP to 2-dim space.}
    \label{fig:SSL_CIFAR_vis}
    \vspace{-1.0em}
\end{figure}
\begin{figure}[b]
    \vspace{-1.0em}
    \centering
    \includegraphics[width=1.0\linewidth]{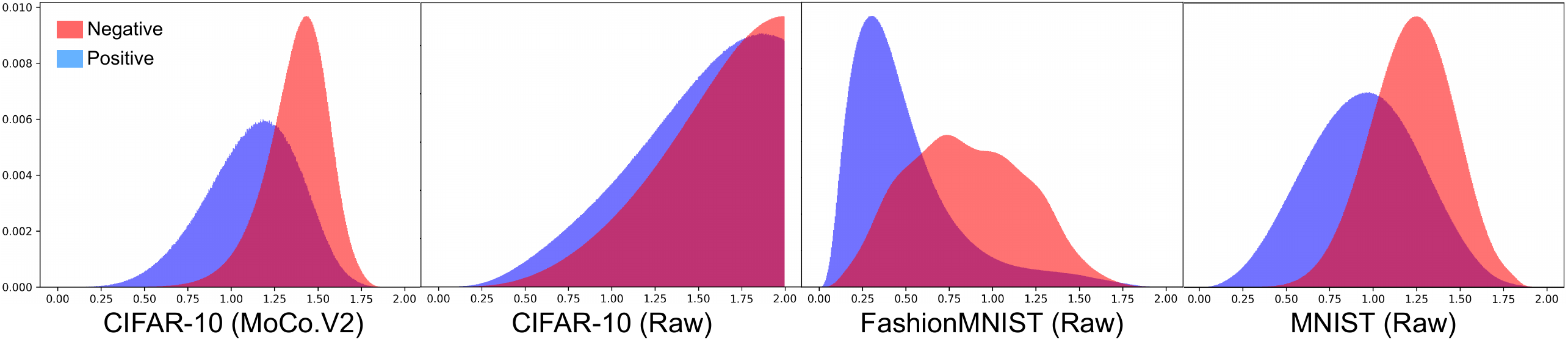}
    \vspace{-2.0em}
    \caption{Histograms of the pair-wise cosine distance of the positive (blue) and negative pairs (red) on STL-10, CIFAR-10, and MNIST datasets. Note that MoCo.v2 and raw denote the distance between the latent space (MoCo.v2) and raw feature spaces.}
    \label{fig:SSL_CIFAR_plot}
    \vspace{-0.5em}
\end{figure}

\paragraph{Relationship between SSL and DR}
We further discuss the relation between SSL and DR tasks (the \textit{dynamic} and \textit{static} $\tilde p_{X}$) with GenURL to explain the desired structures of data in URL tasks. 
Meanwhile, we find that the instance discrimination prior knowledge in SSL tasks is more useful for downstream tasks like clustering and classification of complex scenarios, as shown in Figure~\ref{fig:SSL_CIFAR_vis}. Therefore, we can choose a proper URL task for various scenarios: the DR task is suitable for MNIST and FMNIST datasets where the input spaces are discriminative and reliable, while the SSL task suits more complex datasets like CIFAR-10/100 where the input spaces are unreliable.

\paragraph{Hyperparameters}
Firstly, we compare the effects of hyper-parameters in GenURL for DR and SSL tasks. As shown in Figure~\ref{fig:dr_ablation}, GenURL prefers smaller $\nu_{Z}$, \textit{\textit{i.e.},} using $\nu=0.01$ to balance the local and global structures. 
Figure \red{5} shows that GenURL prefers $\nu_{Z}=100$ for representations with strong discriminative abilities.
Then, we find that the prior knowledge of instance discrimination in SSL tasks is more useful for downstream tasks like clustering and classification of complex scenarios. As shown in Figure~\ref{fig:SSL_CIFAR_vis}, we summarize the learned representations with DR and SSL methods on CIFAR-10 and find that the results of SSL (adopting the instance discrimination prior knowledge) are more reliable and useful to downstream tasks like clustering and classification than DR. Empirically, DR methods are employed on ``simple'' datasets (\textit{e.g.,} MNIST and COIL-20) with reliable geometric structures rather than ``complex'' datasets (\textit{e.g.,} CIFAR-10 and ImageNet) with high-dimensional and redundant features. We hypothesize that this might depend on the property of the dataset. To verify our assumption, we compute the pair-wise distance between raw input samples and the latent space of the SSL model, as shown in Figure~\ref{fig:SSL_CIFAR_plot}, to show the difference between DR and SSL tasks. We find that the input space is discriminative and reliable enough on MNIST and FMNIST for the DR task, while it is unreliable on CIFAR-10.
Therefore, we can conclude as follows: when the dataset is reliable, GenURL can employ the BCE loss and the \textit{static} $\tilde p_{X}$ to perform DR tasks (or GE tasks on the graph data); when the dataset contains high-dimensional redundant features, GenURL can adopt the GKL loss and the \textit{dynamic} $\tilde p_{X}$ with the prior knowledge to conduct SSL tasks.

\paragraph{Complexities}
GenURL has a constant algorithmic complexity in the four URL tasks, thanks to its property of unification. It improves performance without adding extra complexity. Unlike GNN and GCN, the proposed GenURL doesn't require neighborhood aggregation operations, making the complexity agnostic to the network architectures and the kNN graph in the input space. 
In Sec.~\ref{sec:Method}, we build an undirected graph before training and calculate the adjacency matrix $A$ with $O(n^2)$. The pre-computed results will be saved to get $p_{X}(x_i, x_j)$ in Eq.~\ref{eq:pz}. In each iteration, we calculate the latent space pair-wise similarity $p_{X}(z_i, z_j)$ with $O(n^2)$, assuming the batch size is n (\textit{i.e.}, the whole dataset in GE and DR tasks). In CL and KD tasks, we will practically use a mini-batch of $b$ to reduce the computational complexity to $O(b^2)$.

\section{Limitation and Future work}
\label{app:impact}
As for the societal impacts of GenURL, it can be regarded as a unified framework for the unsupervised representation learning (URL) problem that bridges the gap between various methods. The ablation studies of basic hyper-parameters can reflect the relationship between different URL tasks. The core idea of GenURL is to explore intrinsic structures of the data (the raw input space or empirical metric space) and preserve these structures in the latent space, which might inspire some improvements in various URL tasks. For example, the \textit{dynamic} $\widetilde p_{X}$ is similar to the hard negative mining problem in the SSL task~\cite{2021iclrHCL, 2022aaaiunmix}.

As for the limitations of GenURL, we can conclude three aspects: (i) the proposed framework relies on offline hyper-parameter tuning to adapt to new URL tasks, which makes it tough to handle more than two input similarities, (ii) GenURL cannot deal with the case of discrete empirical spaces well, \textit{e.g.}, the SSL tasks, and the \textit{dynamic} $\widetilde p_{X}$ should be improved in the future work, (iii) the performance of GenURL is still limited by negative samples (sensitive to the size of datasets and the batch size). In future work, we plan to tackle the aforementioned limitations and design a dynamic framework that can learn to optimize the data structural modeling and low-dimensional embedding in an end-to-end manner. Moreover, GenURL will be used in more URL application scenarios.

\section{Conclusions}

We propose a simple but efficient similarity-based framework for unsupervised representation learning (URL), called GenURL, that encodes essential structures from the input data and optional prior knowledge. Specifically, we discuss the loss functions for embedding learning in GenURL and proposed binary cross-entropy loss and general Kullback-Leibler divergence loss. Combined with a specific pretext task, we can adapt GenURL to various URL scenarios in a unified manner and achieve state-of-the-art performances, including self-supervised visual representation learning, unsupervised knowledge distillation, graph embeddings, and dimension reduction. Moreover, ablation studies reflect the relationship between data characters and hyper-parameter settings of GenURL.

\section*{Acknowledgement}
This work was supported by the National Key R\&D Program of China (No. 2022ZD0115100), the National Natural Science Foundation of China Project (No. U21A20427), and Project (No. WU2022A009) from the Center of Synthetic Biology and Integrated Bioengineering of Westlake University.
This work was done when Zhiyuan Chen interned at Westlake University.

\nolinenumbers
\bibliographystyle{IEEEtran}
\bibliography{GenURL}




 


\clearpage
\begin{IEEEbiography}[{\includegraphics[width=1in,height=1.25in,clip,keepaspectratio]{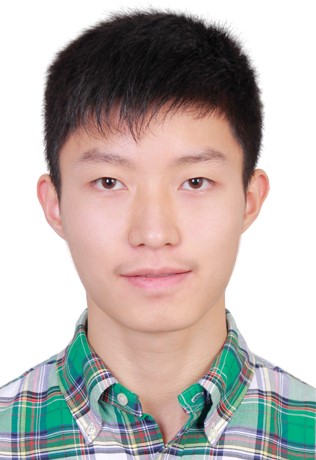}}]{Siyuan Li}
(Student Member, IEEE) received the B.S. degree from the Department of Computer Science and Technology, Nanjing University, Nanjing, China, in 2021. He is currently pursuing the Ph.D. degree in the School of Engineering at Westlake University and the Department of Computer Science and Technology at Zhejiang University, supervised by Prof. Stan Z. Li (Fellow, IEEE).
His main research interests include self-supervised learning, data augmentation, network architecture design in computer vision, and biological application.
\end{IEEEbiography}

\begin{IEEEbiography}[{\includegraphics[width=1in,height=1.25in,clip,keepaspectratio]{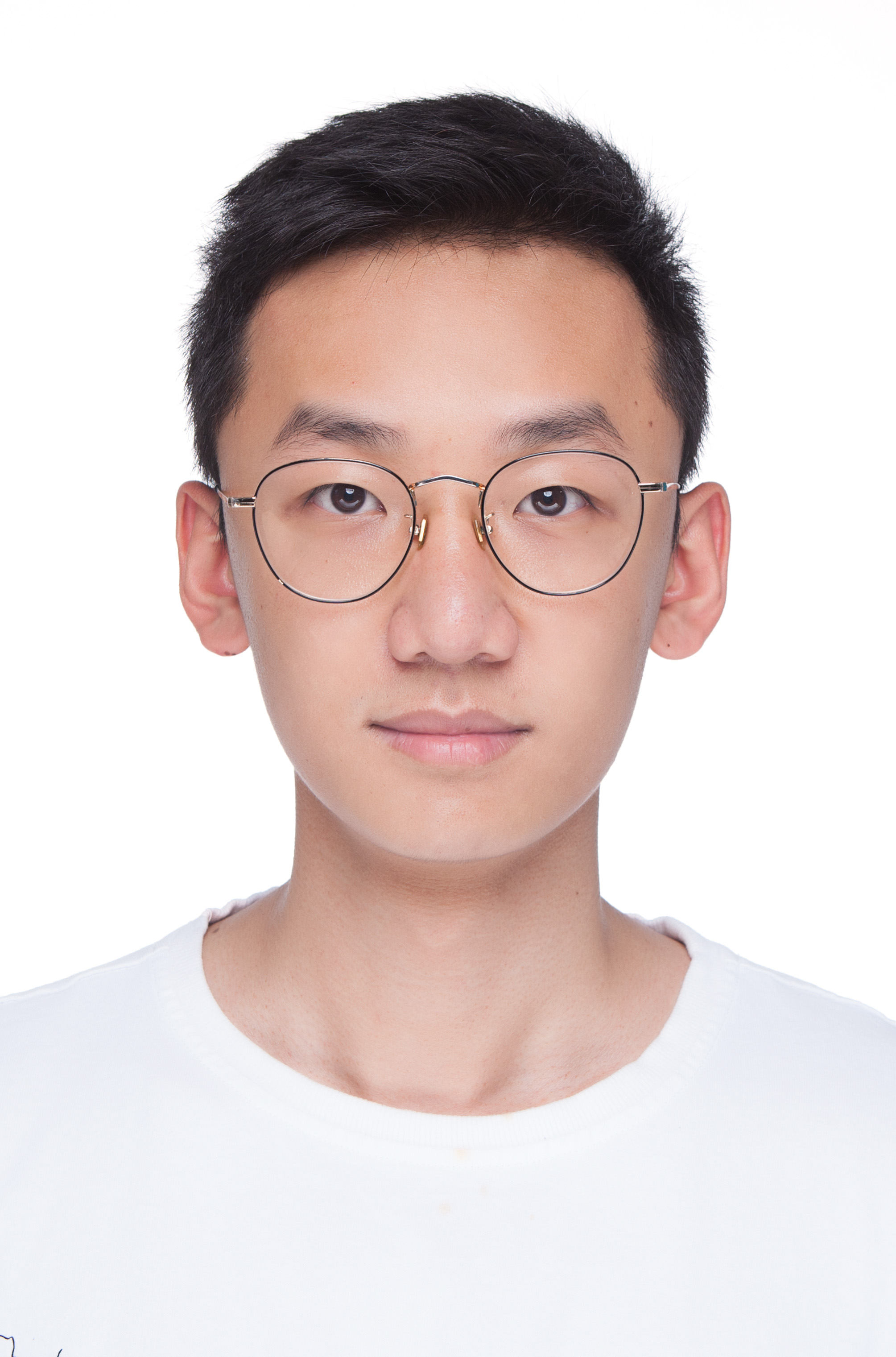}}]{Zicheng Liu}
(Student Member, IEEE) received the B.S. degree from the Department of Information and Computing Science, University of Liverpool, Liverpool, U.K., in 2020. He is currently pursuing the Ph.D. degree with the School of Engineering, Westlake University and Zhejiang University, supervised by Prof. Stan Z. Li (Fellow, IEEE). His research interests include data augmentation, network architecture design, and biological and computer vision applications.
\end{IEEEbiography}

\begin{IEEEbiography}[{\includegraphics[width=1in,height=1.25in,clip,keepaspectratio]{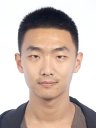}}]{Zelin Zang}
(Student Member, IEEE) received the M.Eng. degree from the Zhejiang University of Technology, Hangzhou, China, in 2020. He is currently pursuing the Ph.D. degree at Westlake University and Zhejiang University, supervised by Prof. Stan Z. Li (Fellow, IEEE).
His research interests include manifold learning, dimension reduction, self-supervised learning in computer vision, and biological applications.
\end{IEEEbiography}

\begin{IEEEbiography}[{\includegraphics[width=1in,height=1.25in,clip,keepaspectratio]{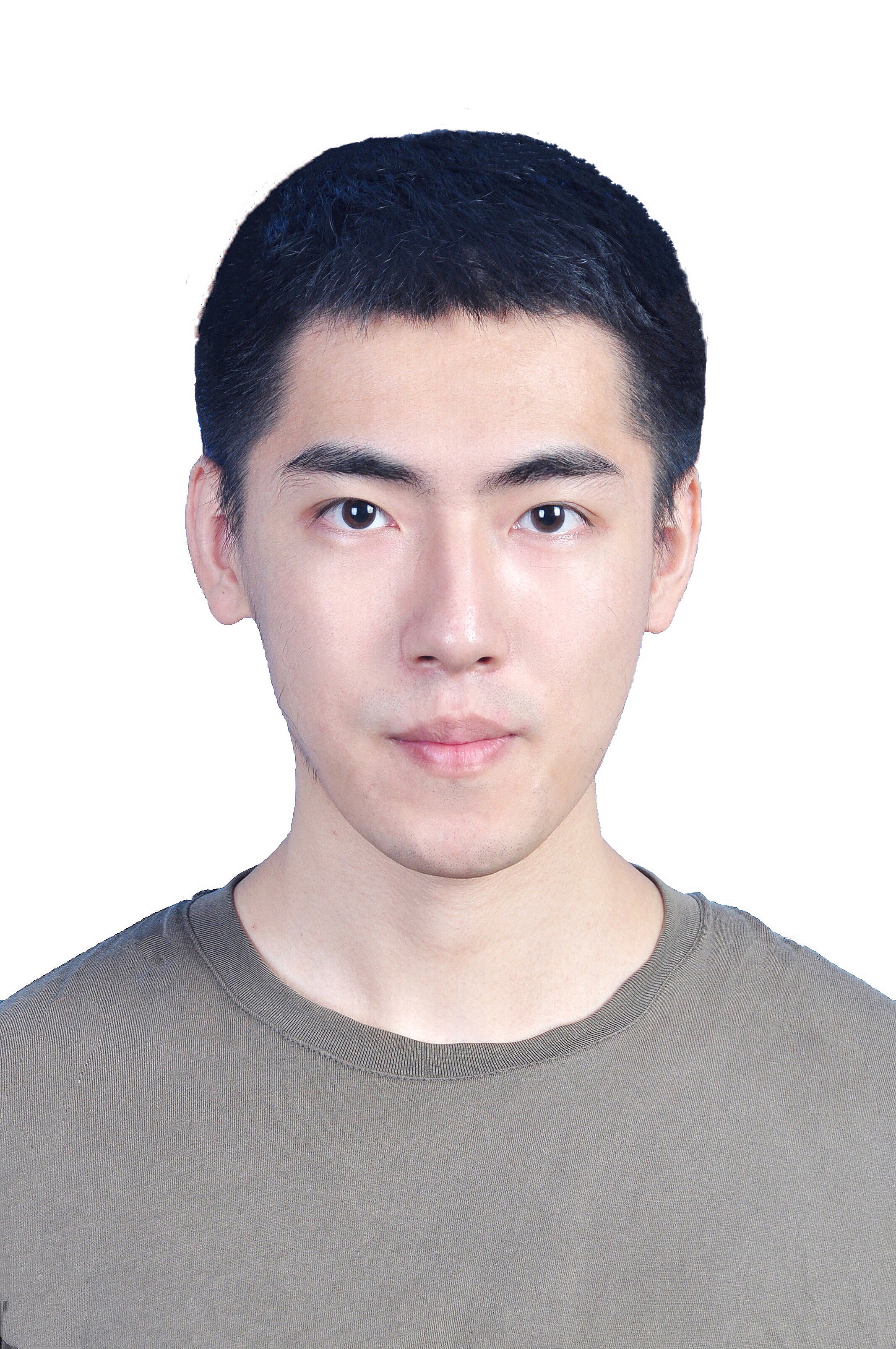}}]{Di Wu}
received the B.S. degree from the Department of Computer Science, Harbin Institute of Technology, Harbin, China, in 2016. He received the M.S. degree in electrical and computer engineering from Boston University, Boston, MA, USA, in 2018. He is currently working toward the Ph.D. degree with the Center of Excellence in Biomedical Research on Advanced Integrated-on-chips Neurotechnologies, School of Engineering, Westlake University, Hangzhou, China. His research interests include self-supervised learning and efficient deep learning for neurophysiological applications.
\end{IEEEbiography}

\begin{IEEEbiography}[{\includegraphics[width=1in,height=1.25in,clip,keepaspectratio]{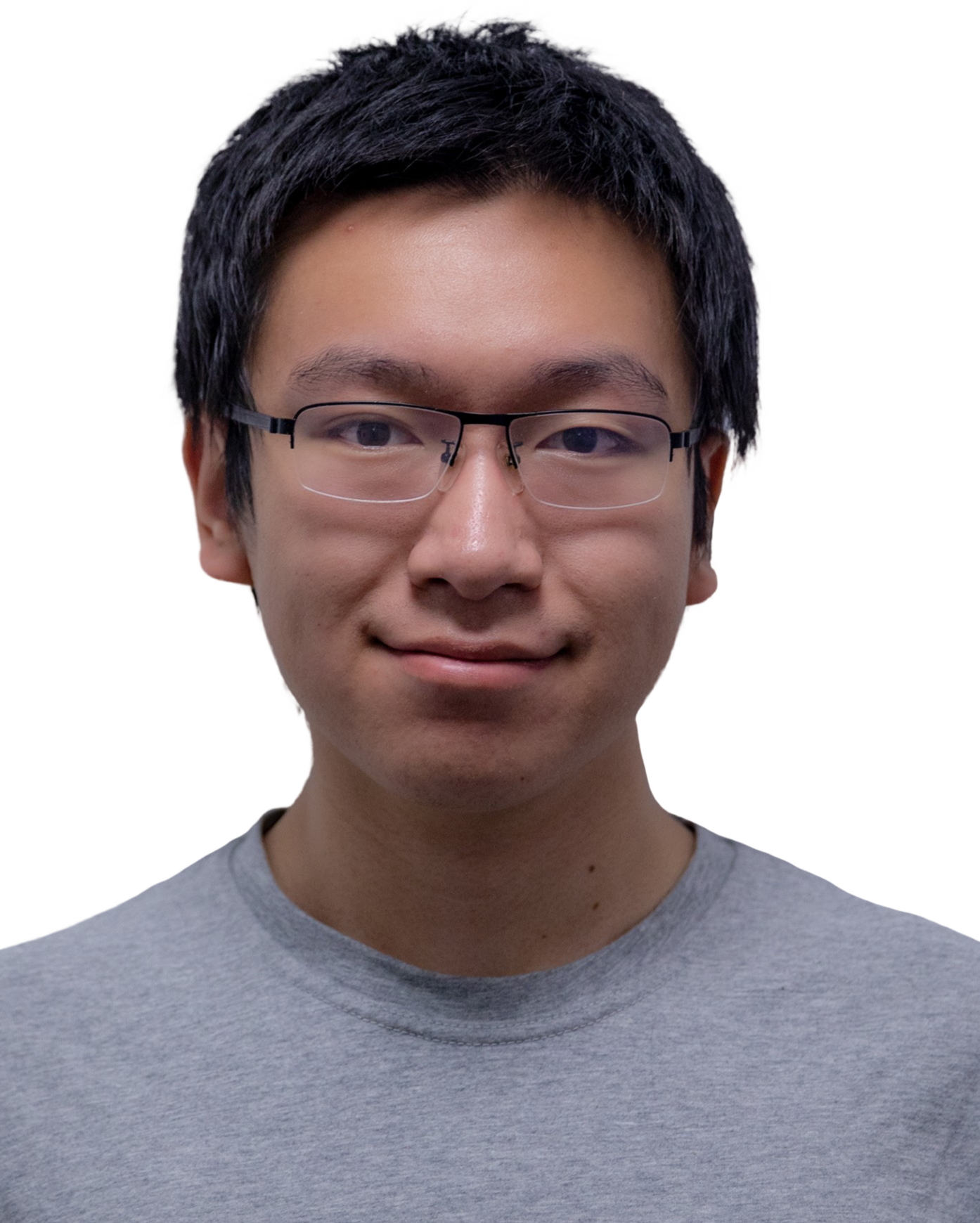}}]{Zhiyuan Chen}
received the B.S. degree from the Australian National University, Australia, in 2022. We worked as a research assistant in the StarBridge program \& a research intern at the Star of Tomorrow program under the guidance of Dr. Pan Deng at Microsoft Research Asia \& Microsoft Research AI for Science from 2021-2022. He is currently a researcher at Deep Potential, Beijing, China, working on macro-molecules. His research interests include AI4Science, computer vision, and deep learning.
\end{IEEEbiography}

\begin{IEEEbiography}[{\includegraphics[width=1in,height=1.25in,clip,keepaspectratio]{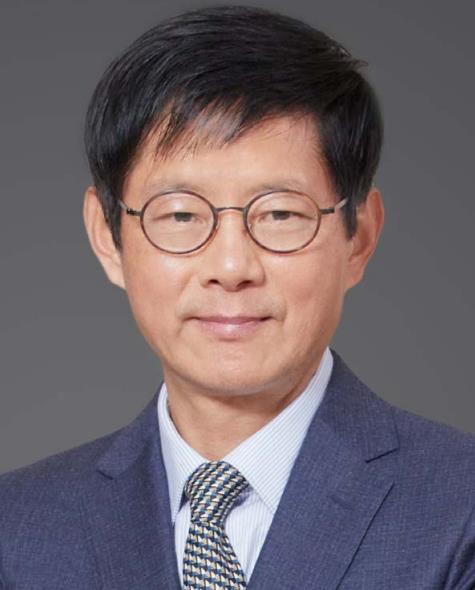}}]{Stan Z. Li}
(Fellow, IEEE) received the B.Eng. degree from Hunan University, Changsha, China, in 1982, the M.Eng. degree from the National University of Defense Technology, Changsha, in 1985, and the Ph.D. degree from the University of Surrey, Guildford, U.K, in 1991.
He was the Director of the Center for Biometrics and Security Research (CBSR), Chinese Academy of Sciences, Beijing, China, from 2004 to 2019. He worked at Microsoft Research Asia, Beijing, as a Research Lead from 2000 to 2004. Prior to that, he was an Associate Professor (Tenure) at Nanyang Technological University, Singapore. He joined Westlake University, Hangzhou, China, as a Chair Professor of artificial intelligence in February 2019. He has published over 400 articles in international journals and conferences and authored and edited 10 books, with over 50,000 Google Scholar citations. Among these are Markov Random Field Models in Image Analysis (Springer), Handbook of Face Recognition(Springer), and Encyclopedia of Biometrics (Springer). His research interests include fundamental research in machine learning, data science, and applied research in multiple AI-related interdisciplinary fields (computer vision, smart sensors, life science, material science, and environmental science).
Dr. Li served as an Associate Editor for IEEE TRANSACTIONS ON PATTERN ANALYSIS AND MACHINE INTELLIGENCE and organized more than 100 international conferences or workshops.
\end{IEEEbiography}



\end{document}